\documentclass[acmtog]{acmart}
\acmSubmissionID{347}

\usepackage{booktabs} 
\usepackage{gensymb}
\usepackage[noend]{algorithmic}
\usepackage[ruled]{algorithm2e} 

\SetAlFnt{\small}
\SetAlCapFnt{\small}
\SetAlCapNameFnt{\small}
\SetAlCapHSkip{0pt}
\newcommand{\name}{SNeRF }
\newcommand{\norm}[1]{\left\lVert#1\right\rVert}

\newcommand{\new}[1]{\textcolor{black}{{#1}}}
\newcommand{\V}[1]{{\bf #1}}
\newcommand{\old}[1]{}
\acmJournal{TOG}

\setcopyright{rightsretained}
\acmJournal{TOG}
\acmYear{2022}\acmVolume{41}\acmNumber{4}\acmArticle{142}\acmMonth{7} \acmDOI{10.1145/3528223.3530107}

\begin{document}
\title{SNeRF: Stylized Neural Implicit Representations for 3D Scenes}

\author{Thu Nguyen-Phuoc}
\email{thunp@fb.com}
\author{Feng Liu}
\email{fliu@cs.pdx.edu}
\author{Lei Xiao}
\email{lei.xiao@fb.com}
\affiliation{%
\institution{Reality Labs Research, Meta}
\country{USA}
}

\begin{teaserfigure}\centering
	\includegraphics[width=\linewidth]{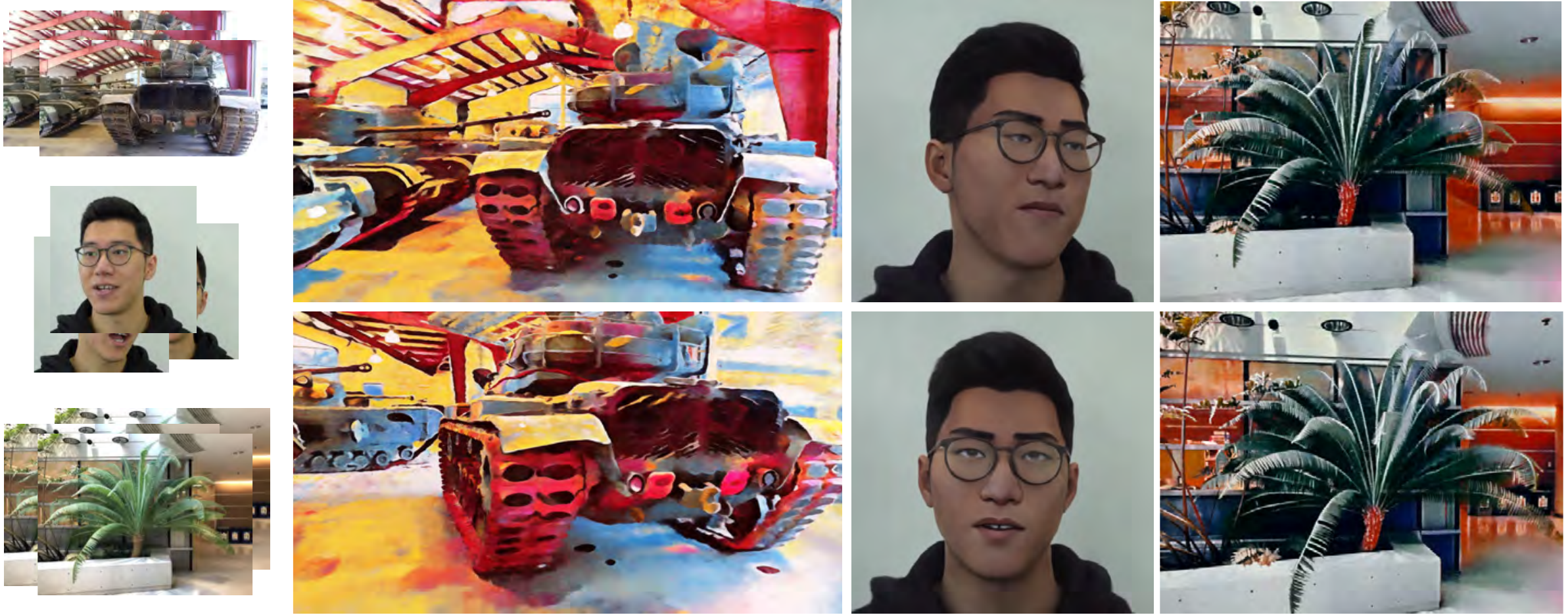}%
	\caption{Given a neural implicit scene representation trained with multiple views of a scene, \name stylizes the 3D scene to match a reference style. \name works with a variety of scene types (indoor, outdoor, 4D dynamic avatar) and generates novel views with cross-view consistency. \label{fig:teaser}}
\end{teaserfigure}

\begin{abstract}

This paper presents a stylized novel view synthesis method.
Applying state-of-the-art stylization methods to novel views frame by frame often causes jittering artifacts due to the lack of cross-view consistency.
Therefore, this paper investigates 3D scene stylization that provides a strong inductive bias for consistent novel view synthesis.
Specifically, we adopt the emerging neural radiance fields (NeRF) as our choice of 3D scene representation for their capability to render high-quality novel views for a variety of scenes. 
However, as rendering a novel view from a NeRF requires a large number of samples, training a stylized NeRF requires a large amount of GPU memory that goes beyond an off-the-shelf GPU capacity. 
We introduce a new training method to address this problem by alternating the NeRF and stylization optimization steps. 
Such a method enables us to make full use of our hardware memory capacity to both generate images at higher resolution and adopt more expressive image style transfer methods.
Our experiments show that our method produces stylized NeRFs for a wide range of content, including indoor, outdoor and dynamic scenes, and synthesizes high-quality novel views with cross-view consistency.

\end{abstract}


\maketitle

\section{Introduction}
With the increasing availability of new social media platforms and display devices, there has been a growing demand for new visual 3D content, ranging from games and movies to applications for virtual reality (VR) and mixed reality (MR). In this paper, we focus on the problem of stylizing 3D scenes to match a reference style image. Imagine putting on a VR headset and walking around a 3D scene: one is no longer constrained by the look of the real world, but instead can view how the world would look like through the artistic lenses of Pablo Picasso or Claude Monet.

Naively applying image-based stylization techniques \cite{neuralStyleTransfer} to 3D scenes might lead to flickering artefacts between different views, since each view is stylized independently without any consideration for the underlying 3D structure. 
Therefore, recent work has explored various choices of 3D representations to address this issue: one can stylize the underlying 3D scenes and then render new consistent views from them \cite{KPLD21, huang_2021_3d_scene_stylization}. However, these methods does not capture the target style well since they only stylize the scene's appearance, although geometry is also an important part of styles~\cite{Kim20DST, 9577906}.

Recently, neural radiance fields (NeRF) \cite{mildenhall2020nerf} offers a compact 3D scene representation that produces high-quality novel-view synthesis results. Later work shows the flexibility of NeRF as 3D scene representations, ranging from large outdoor scenes \cite{kaizhang2020} to dynamic avatars \cite{Gafni_2021_CVPR}. Its compactness, expressiveness and flexibility make NeRF an attractive choice of 3D representation for stylization.
However, adopting NeRF for neural style transfer poses a great memory constraint. 
To render a pixel from NeRF, one has to sample densely along a camera ray. This requires high memory usage for rendering and performing back-propagation (for example, it takes 17, 934 MB to render an image patch of size 81 $\times$ 67 \cite{chiang2021stylizing}).
Concurrent work by \citet{chiang2021stylizing} addresses this limitation by performing stylization on rendered patches of NeRF instead of the whole images.
However, results from patch-based approaches tend to suffer from global style inconsistencies due to the mismatch between patches in the target style and content images~\cite{huang2017adain}.

We propose to combine NeRF and image-based neural style transfer to perform 3D scene stylization. 
While NeRF provides a strong inductive bias to maintain multi-view consistency, neural style transfer enables a flexible stylization approach that does not require dedicated example inputs from professional artists \cite{Stylit, styblit, Texler20-SIG}. 
Additionally, we address the memory limitations of NeRF by splitting the 3D scene style transfer process into two steps that run alternatingly. 
This enables us to fully utilize the memory capacity of our hardware to either render NeRF or perform neural style transfer on images with high resolutions. 

In this paper, we present SNeRF, a 3D scene neural stylization framework that generates novel views of a stylised 3D scene while maintaining cross-view consistency. Our primary technical contributions include the following:
\begin{itemize}
    \item We introduce a novel style transfer algorithm with neural {\emph{implicit}} 3D scene representations, producing high-quality results with cross-view consistency.
    \item We introduce a general, plug-and-play framework, where various implicit scene representations and stylization methods can be plugged in as a sub-module, enabling results on a variety of scenes: indoor scenes, outdoor scenes and 4D dynamic avatars.
    \item We develop a novel training scheme to effectively reduce the GPU memory requirement during training, enabling high-resolution results on a single modern GPU.
    \item Through both objective and subjective evaluations, we demonstrate that our method delivers better image and video quality than state-of-the-art methods.
\end{itemize}
\section{Related work}
\subsection{Image and video style transfer}
Style transfer aims to synthesize an output image that matches a given content image and a reference style image. 
Image analogies by Hertzmann {\emph{et al.}}~\shortcite{hertzmann2001image} and follow-up work \cite{deepImageAnalogies} address this problem by finding semantically-meaningful dense correspondences between the input images, which allows effective visual attribute transfer. 
However, they require the content and style image to be semantically similar.
Stylize-by-example approaches also adopt a patch-based approach using high-quality examples as guidance \cite{Stylit, Sykora19-EG, Texler19-NPAR}.
Despite impressive results, these methods require dedicated guiding examples provided by professional artists.
A blind approach to image stylization, neural style transfer, is later on proposed by \citet{neuralStyleTransfer}.
Unlike example-based approaches, neural style transfer can perform stylization on arbitrary style reference images. 
Originally, this is done by optimizing the output image to match the statistics of the content and style images, which are computed using a pre-trained deep network.
This optimization process is later replaced by feed-forward networks to speed up the stylization process \cite{Johnson2016Perceptual, instanceNorm}. 
Instead of redoing the stylization for every new style, recent frameworks use the adaptive instance normalization (AdaIN) \cite{huang2017adain}, whitening and coloring transform (WCT)~\cite{WCT-NIPS-2017}, linear transformation (LST)~\cite{li2018learning}, or feature alignment~\cite{svoboda2020twostage} to perform style transfer with arbitrary new styles at test time.

While relatively similar to image style transfer, video style transfer methods mainly focus on addressing the temporal consistency across the video footage.
Recently, key-frame based approaches \cite{Jamriska19-SIG, chiang2021stylizing} expand stylize-by-examples to videos and have shown impressive results, but require guiding examples from artists for every keyframe.
For blind approaches without dedicated guiding style reference, this can be done using optical flow to calculate temporal losses \cite{opticalFlowdistillation, 8237388} or align intermediate feature representations \cite{GaoACCV, 8100228} to stabilize models' prediction across nearby video frames. 
Recently, there have been efforts to improve consistency and speed for video style transfer for arbitrary styles through temporal regularization \cite{ReReVST2020}, multi-channel correlation \cite{Deng_Tang_Dong_Huang_Ma_Xu_2021}, and bilateral learning \cite{9423312}. 
\new{Similarly, style transfer for stereo images \cite{Chen_2018_CVPR, Gong_2018_ECCV} also aims to achieve cross-view consistency by using dense pixel correspondences (via stereo matching) constraints.
However, these methods mostly focus on improving short-range consistency between nearby frames or views, and do not support novel view synthesis.
}
\subsection{3D style transfer}
While style transfer in the image domain is a popular and widely studied task, style transfer in the 3D domain remains relatively new. 
Most approaches focus either on stylizing a single object, using either meshes \cite{Ma:2014:AST} or point clouds \cite{segu20203dsnet}, or material \cite{nguyen:2012:3DMaterialStyle}.
Later work focuses on performing style transfer on both geometry and texture \cite{kato2018renderer, yin2021_3DStyleNet, Hauptfleisch20-PG}, but still limited to single objects.

For style transfer at 3D scene level, recent approaches use point clouds \cite{Cao_2020_WACV, huang_2021_3d_scene_stylization, KPLD21} or meshes \cite{stylemesh} as scene representations. 
However, these approaches are limited to static scenes.
Concurrent work by \citet{chiang2021stylizing} uses implicit scene representations, in particular, NeRF, for 3D scene stylization.
However, they only work with static outdoor scenes, while we show that our method works on a variety of scene types. 
Moreover, due to memory constraints, they only perform stylization on image patches, which tend to generate results that lack global style consistency and thus does not capture the reference style well.
Meanwhile, our proposed stylization approach can be trained with images in full resolution and adopt more memory-intensive style transfer approaches such as ArcaneGAN \cite{arcanegan}.
Finally, their method only focuses on stylizing the appearance of the scene, although geometry has been acknowledged to be an important factor of style \cite{Kim20DST, 9577906}. 
\subsection{Novel view synthesis}
Novel view synthesis aims to estimate images at unseen viewpoints from a set of posed source images.
When source images can be sampled densely, light field approaches work well~\cite{Gortler:96,Levoy:96}. Other approaches often use the scene geometrical proxy to warp and blend input views to create novel views~\cite{buehler2001unstructured,Chaurasia:13,zitnick2004high,penner2017soft}.
\begin{figure*}
	\includegraphics[width=0.7\linewidth]{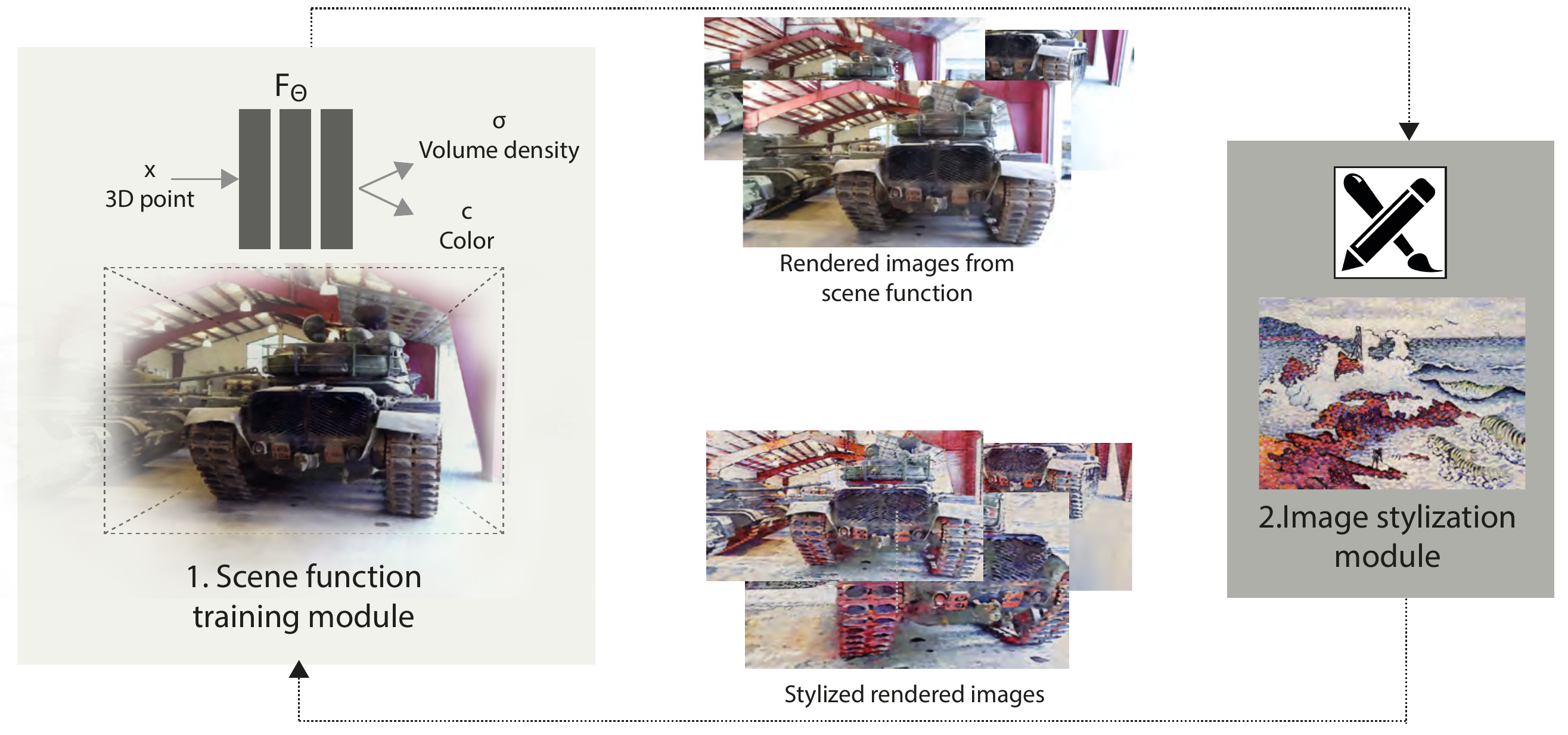}%
	\caption{\label{fig:diagram} \textbf{Overview}: We propose an alternating training approach to stylize implicit scene representations. For one iteration: (1) Given a pre-trained scene function, we render images from different views. (2) We then stylize these images using the image stylization module. (3) We train the scene function to match multi-view stylized images similar to training a NeRF function. In the next iteration, (1) we again render images from different views from a now more stylized scene function, (2) perform image stylization on this new set of images, and (3) train NeRF with the new set of images.}
\end{figure*}
Recently, a wide variety of deep learning-based approaches have been developed for novel view synthesis~\cite{flynn2016deepstereo,Kalantari16,Hedman18}. Aliev~\emph{et al.}~\shortcite{aliev2020neural} develop a neural point-based rendering method. This method associates a feature descriptor for each 3D point, projects the feature descriptors to the target view, and finally uses a neural network to synthesize the target view.
Deferred neural rendering employs a similar approach to mesh-based rendering~\cite{thies2019deferred}. 
Sitzmann~\emph{et al.}~\shortcite{sitzmann2019deepvoxels} develop a scene representation, called DeepVoxels, that can encode the view-dependent effects without explicitly modeling the scene geometry. 
Zhou~\emph{et al.}~\shortcite{zhou2018stereo} estimate multi-plane images from two input images as the scene representation, which can be projected to the target view to render the target view. 
This method is further improved to take more input views~\cite{mildenhall2019local}, handle larger viewpoint shifts~\cite{srinivasan2019pushing}, and produce VR videos~\cite{Broxton2020}. 
Mildenhall~\emph{et al.}~\shortcite{mildenhall2020nerf} represent scenes as neural radiance fields (NeRF), and show impressive results for novel view synthesis. Since then, a large number of NeRF methods have been developed~\cite{kaizhang2020,martinbrualla2020nerfw,li2021neural,barron2021mipnerf}, which are described in a survey by \citet{dellaert2021neural}. 
Given NeRF's ability to render high-quality views and represent a variety of scene types, we adopt NeRF for stylization to render cross-view consistent stylized novel views.

\section{Method}
Given a 3D scene, we aim to manipulate it such that rendered images of this scene match the style from a reference image $I_{style}$.
Additionally, rendered images of the same scene from different views should be consistent. 
In this work, we use NeRF as our choice of scene representation for its compactness and flexibility.
We propose a memory-efficient training approach that alternates between stylization and NeRF training.
This enables us to make full use of our hardware memory for either stylization or training NeRF, but not both at the same time, and thus achieve results with high resolution. 
%
%
\subsection{Preliminaries}
\subsubsection{NeRF overview}\label{sec:nerf}
NeRF is a continuous 5D function whose input is a 3D location $\V{x}$ and 2D viewing direction $\V{d}$, and whose output is an emitted color $\V{c} = (r, g, b)$ and volume density $\sigma$.
NeRF is approximated by a multi-layer perceptron (MLP): $F_{\Theta}:(\V{x}, \V{d}) \mapsto (\V{c}, \sigma)$, which is trained using the following loss function:
\begin{equation}
\label{loss_nerf}
L_{NeRF}(F_{\Theta}) = \frac{1}{M} \sum^M_{i=1}\norm{\V{c}(\V{r}_i) - \V{c}'(\V{r}_i)}_2
\end{equation}
where $\{\V{r}_i\}_{i=1}^M$ is a batch of randomly sampled camera rays using the corresponding camera poses and the camera intrinsic at each optimization step, $\V{c'}(\V{r}_i)$ is the color of a pixel rendered from $F_{\Theta}$, and $\V{c}(\V{r}_i)$ is the ground truth pixel color. 

In this work, we assume that the stylization process starts with a NeRF pre-trained with realistic RGB images $\{x_i\}^N_{i=1}$ and corresponding camera poses $\{\theta_i\}^N_{i=1}$. 
We apply our approach to 3 different NeRF scene types: classic NeRF \cite{mildenhall2020nerf} for indoor scenes, NeRF++ \cite{kaizhang2020} for 360\degree outdoor scenes and finally, dynamic 4D human avatar \cite{Gafni_2021_CVPR}.

\subsubsection{Style transfer overview}
Given a content image $I_{content}$ and target style reference image $I_{style}$, we want to generate a new image $x'$ that matches the style of $I_{style}$, but still maintain the content of $I_{content}$.
This is done by optimizing the generated image to match the content statistics of the content image and the style statistics of the style reference image using the following loss functions:
\begin{align}
L_{Transfer} &= L_{Content}(I_{Content}, x') + L_{Style}(I_{Style}, x') \\
L_{Content} &= \norm{\Phi(I_{Content}) - \Phi(x')}_2 \\
L_{Style} &= \norm{\Phi(I_{Style}) - \Phi(x')}_2  
\end{align}
where $\Phi$ are the image statistics, which are usually features extracted from different layers of a pre-trained network such as VGG \cite{VGG}.
In our case, $I_{content}$ is a rendered view of the scene function, and $x'$ is a stylized version of that view.
While most of the results in this paper is stylized using the neural style transfer algorithm proposed by \citet{neuralStyleTransfer}, we also use a GAN-based method \cite{arcanegan} to stylize dynamic 4D avatars. 

\subsection{Stylizing implicit scene representation}
We stylize a 3D scene represented as a NeRF to match a reference style image $I_{style}$ using the following loss function:
\begin{equation}
    L_{\name} = L_{NeRF} + L_{Transfer} 
\end{equation}
where $L_{Transfer}$ performs stylization to match a given style image and $L_{NeRF}$ maintains the underlying scene structure to preserve multi-view consistency.

Previous work \cite{huang_2021_3d_scene_stylization, chiang2021stylizing, KPLD21} optimizes for both losses at the same time to perform scene stylization.
This would require rendering full images (or patches as proposed by \citet{chiang2021stylizing}) from NeRF to compute $L_{Transfer}$ at \textit{every training step}, which is time consuming.
Additionally, this approach requires that the memory has to be shared between three memory-intensive components: the feature extractor (such as VGG) to compute image statistics, the volumetric renderer of NeRF, and back-propagation.
This greatly limits the resolution of stylized results as well as the choice of stylization methods.
\new{For example, with a 32GB GPU (NVIDIA V100), we could only perform stylization simultaneously for images at size 252 $\times$ 189 using VGG-16-based losses  similar to \cite{neuralStyleTransfer}, and quickly ran into OOM error with larger images at size 366 $\times$ 252.}

To address the memory burden of the methods described above, we propose an alternating training regime inspired by coordinate descent.
Our insight is that we can decouple $L_{Transfer}$ and $L_{NeRF}$, and minimize one at a time.
To compute $L_{Transfer}$, we only need the feature extractor, the target style image, and rendered images of the scene, which can be precomputed from NeRF.
Meanwhile, to compute $L_{NeRF}$, we only need the volumetric renderer and target images, which can be precomputed by a separate stylization process.
In practice, we train NeRF with $L_{NeRF}$ for a number of steps on batches of randomly sampled rays across different views, before rendering a set of images at different views to compute $L_{Transfer}$.
Figure \ref{fig:diagram} provides an overview of our method.

The proposed alternating training regime allows one to dedicate the full hardware capacity to either image stylization or NeRF training. 
For image stylization, this enables us to perform stylization on the whole image and achieve more globally consistent stylized results.
For NeRF training, we can train NeRF to generate results at higher resolution, and apply our method to dynamic scenes (in particular, dynamic avatar).
\new{With the same hardware, our training regime can now stylize NeRF to synthesize images at size 1008 $\times$ 756, 4 times larger than what we previously could when performing training and stylization simultaneously.}
This also opens up potentials to use more expressive pre-trained models to compute $L_{Transfer}$, such as StyleGAN \cite{karras2019} or CLIP \cite{CLIP}.
\subsection {Alternating training regime details}
Starting from a set of ``realistic'' RGB image $\{x_i\}^N_{i=1}$ rendered from pre-trained NeRF using camera poses $\{\theta_i\}^N_{i=1}$, we perform style transfer independently on each image (as the target content) by minimizing $L_{Transfer}$. 
Note that after this step, the stylized images are not necessarily multi-view consistent.
Secondly, we use this set of stylized images $\{{x'}_i\}^N_{i=1}$ as target images to train the NeRF scene function $F_{\Theta}$ using $L_{NeRF}$.
Note that here we can train NeRF on batches of random rays across multiple views, instead using full images which can be time and memory-consuming.
Finally, using the stylised NeRF, we render a new set of images.
While these images might not yet capture the full details of the target style, they are multi-view consistent thanks to the underlying scene structure of NeRF.
In the next iteration, we perform stylization on the new set of (more stylized) images of NeRF. 
By boostraping the image stylization algorithm to the output images of NeRF, we obtain more multi-view consistent stylized results, even when each view is stylized independently.
We then use the new set of stylized images to further finetune NeRF.
The outline of the overall stylization process is described in Algorithm~\ref{algo:Training}. Please refer to the supplementary video for converging results at each iteration. 

\begin{algorithm}
\caption{Neural Implicit Scene Representation Stylization}
\label{algo:Training}
\KwIn{Neural implicit scene function $F_{\Theta}$ pre-trained on realistic multi-view images, target style image $I_{Style}$}
\KwOut{Stylised implicit scene function $\hat{F}_{\Theta}$.}
\begin{algorithmic}
\STATE{Initialize $\hat{F}_{\Theta}$ with ${F}_{\Theta}$.}
\FOR{each iteration t = 1,...,T}
\STATE{Render a set of images $\{x_i\}^K_{i=1}$ using the stylized scene function $\hat{F}_{\Theta}$.}
\STATE{Optimize the stylized images $\{x_i'\}^K_{i=1}$ to minimize the style transfer loss: $\sum^K_{i=1}{L_Style}(I_{Style}, x_i') + L_{Content}(x_i, x_i')$.}
\STATE{Optimize $\hat{F}_{\Theta}$ to minimize $L_{NeRF}(\hat{F}_{\Theta})$ using $\{x_i'\}^K_{i=1}$ as reference.}
\ENDFOR
\end{algorithmic}
\end{algorithm}

\subsection{Implementation Details}
For stylization, we use a pre-trained VGG\-16 network \cite{VGG} to extract the image statistics.
In particular, we use layer \textit{relu4\_1} to extract image features for the content loss, and layers \textit{relu1\_1}, \textit{relu2\_1}, \textit{relu3\_1} and \textit{relu4\_1} for the style loss.
For the 4D avatar, we use a pre-trained ArcaneGAN model \cite{arcanegan}.

For each scene, we perform scene stylization for 5 iterations ($T=5$ in Algorithm \ref{algo:Training}).
For each iteration, we perform neural style transfer optimization for 500 steps for each input image, and train the scene function NeRF for 50000 steps (100000 steps for NeRF++). We use the learning rate of 5e-4 for all of our experiments. 
We train each model using one NVIDIA V100 GPU.

\section{Results}
To show the flexibility of our stylization method, we train \name on 3 different scene types: indoor scenes, outdoor scenes and dynamic avatar. For indoor scenes, we train NeRF using scenes \textit{Fern} and \textit{TRex} from the LLFF dataset \cite{mildenhall2019llff}. For outdoor scenes, we train NeRF++ \cite{kaizhang2020} using scenes \textit{Truck}, \textit{Train}, \textit{M60} and \textit{Playground} from the Tank and Temples dataset \cite{TankAndTemple}. We also use our method to stylize 4D avatars \cite{Gafni_2021_CVPR}.
For NeRF scenes, we stylize 3D scenes using images at size 1008 $\times$ 756. 
For NeRF++, we use images at 980 $\times$ 546 for \textit{Truck}, 982 $\times$ 546 for \textit{Train}, 1077 $\times$ 546 for \textit{M60} and 1008 $\times$ 548 for \textit{Playground}. 
For the dynamic 4D avatar, we train with images at size 512 $\times$ 512.
\new{We use all available training views for image stylization.}

In Section \ref{sec:qualitative_results} and \ref{sec:quantitive_results}, we compare \name with both 2D and 3D approaches. 
In particular, we compare our method to the following 4 categories of methods:
\begin{itemize}
    \item {Image stylization $\rightarrow$ Novel view synthesis: we perform image stylization on the input images and synthesize new views from them using LLFF \cite{mildenhall2019llff}.}
    \item {Novel view synthesis $\rightarrow$ Image stylization: we perform novel view synthesis using the input images and then stylize each new view independently using AdaIN \cite{huang2017adain}, WCT \cite{WCT-NIPS-2017} and LST \cite{li2018learning}.}
    \item {Novel view synthesis $\rightarrow$ Video stylization: we perform novel view synthesis using the input images, compile the results into a video, and then perform video stylization using ReReVST \cite{ReReVST2020} and MCCNet \cite{Deng_Tang_Dong_Huang_Ma_Xu_2021}.}
    \new{\item{3D scene stylization $\rightarrow$ Novel view synthesis: we compare our method with StyleScene by \citet{huang_2021_3d_scene_stylization}, a point cloud-based approach and with \citet{chiang2021stylizing}, a NeRF-based approach but with a patch-based stylization strategy.}}
\end{itemize}
\subsection{Qualitative results}
 \label{sec:qualitative_results}

\begin{figure*}
	\includegraphics[width=\linewidth]{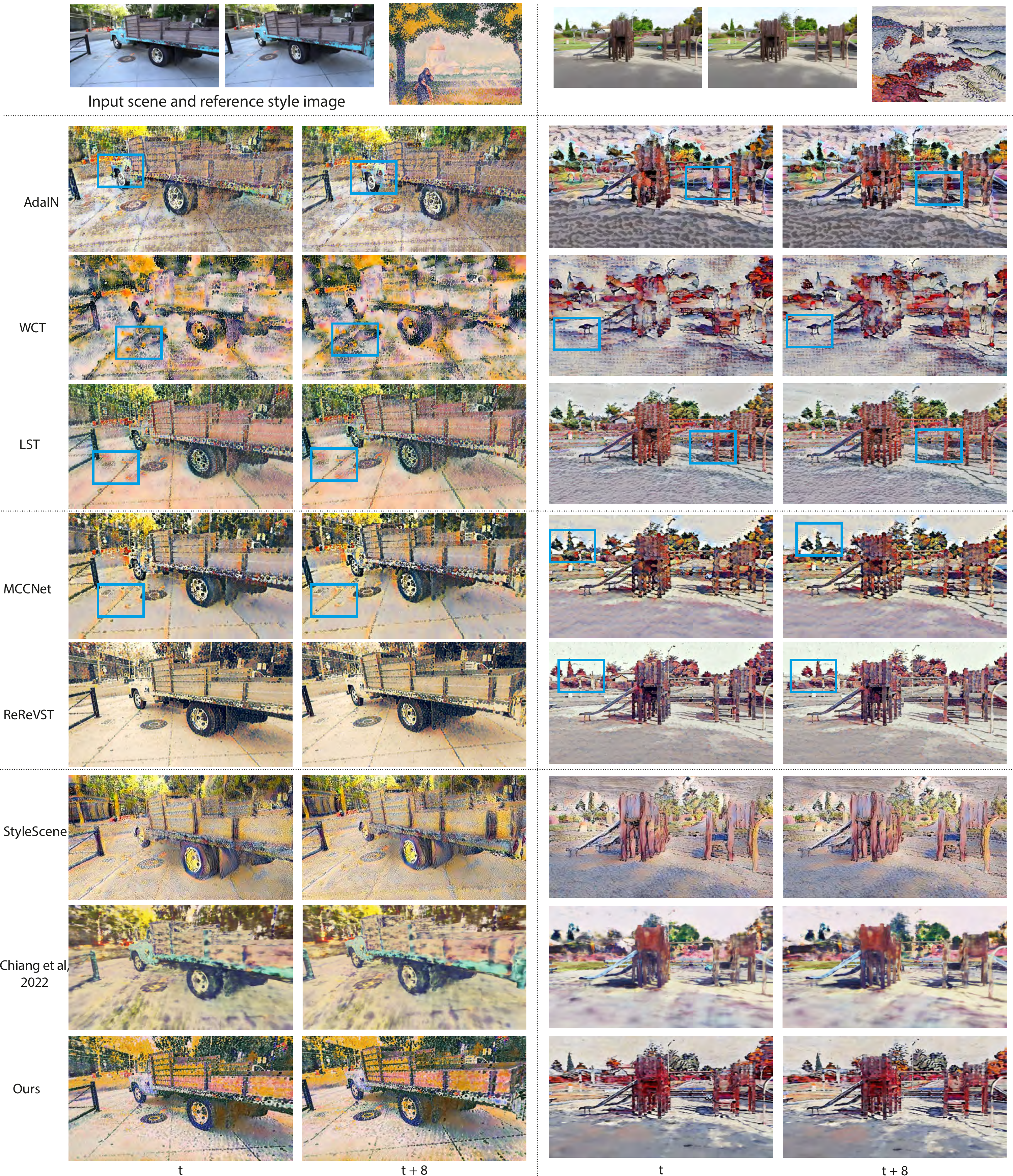}
	\caption{\label{fig:qualitative_comparison}
		\textbf{Qualitative comparisons}. \new{We encourage readers to look at the supplementary material to compare the consistency between different methods. We show stylized results from two frames that are far apart ($t^{th}$ and $(t + 8)^{th}$ frame). Cross-view inconsistencies are highlighted in blue boxes.}
	}
\end{figure*}
\begin{figure*}
	\includegraphics[width=\linewidth]{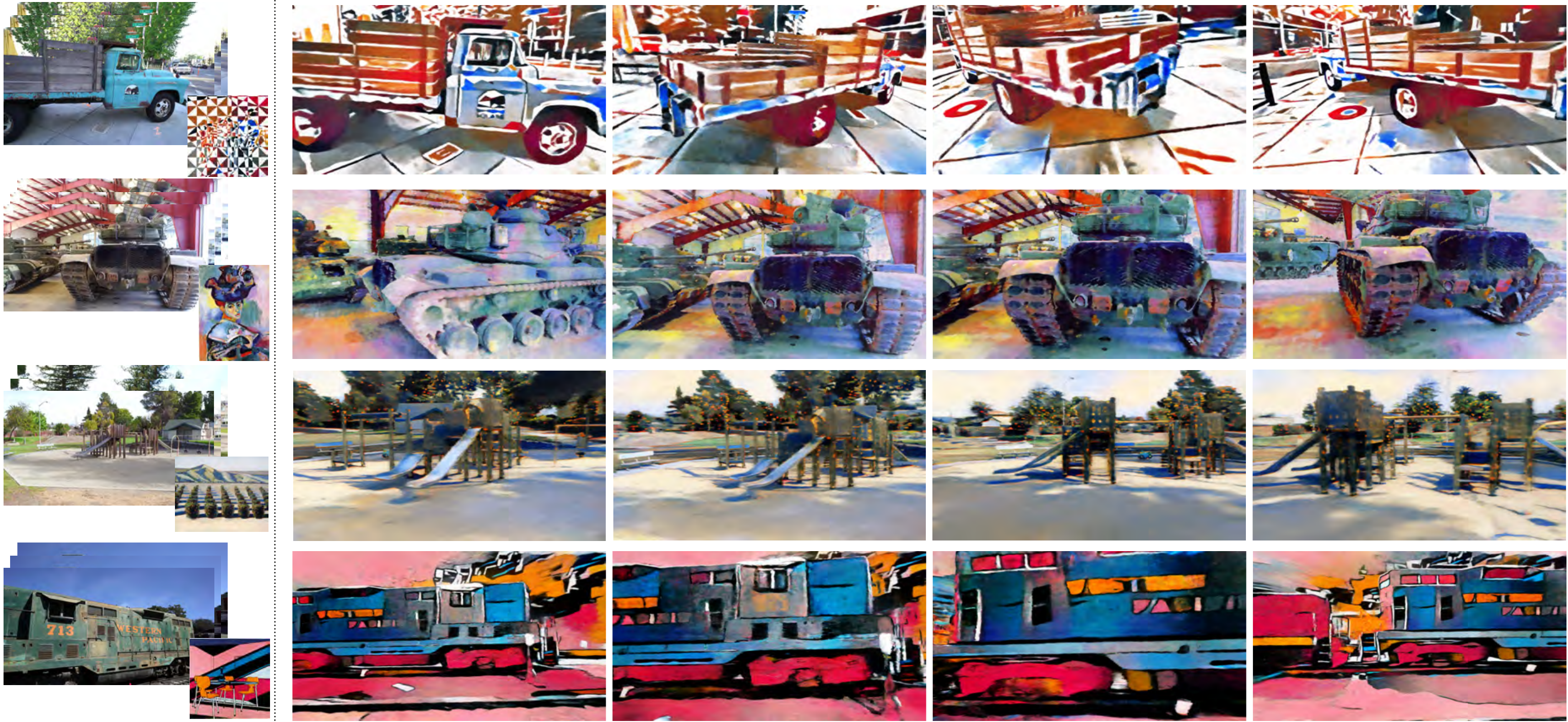}%
	\caption{\label{fig:qualitative_results_DDP}
		SNeRF's stylization results on 360\degree scenes using NeRF++. Here we show our results from different views.
	}
\end{figure*}
\begin{figure*}
	\includegraphics[width=\linewidth]{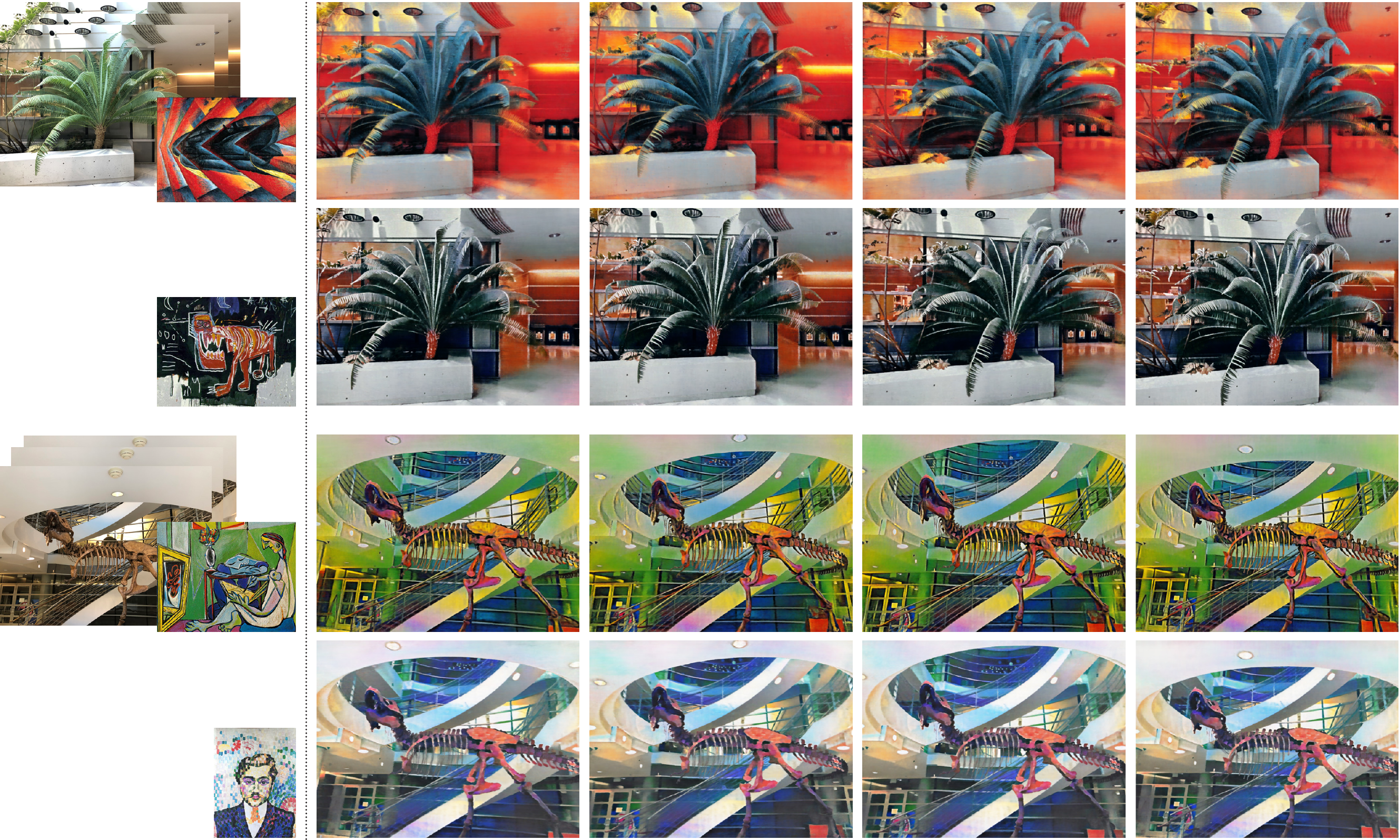}%
	\caption{\label{fig:qualitative_results}
		SNeRF's stylization results on indoor scenes using NeRF. Here we show our stylization results from different views.
	}
\end{figure*}
\begin{figure*}
	\includegraphics[width=\linewidth]{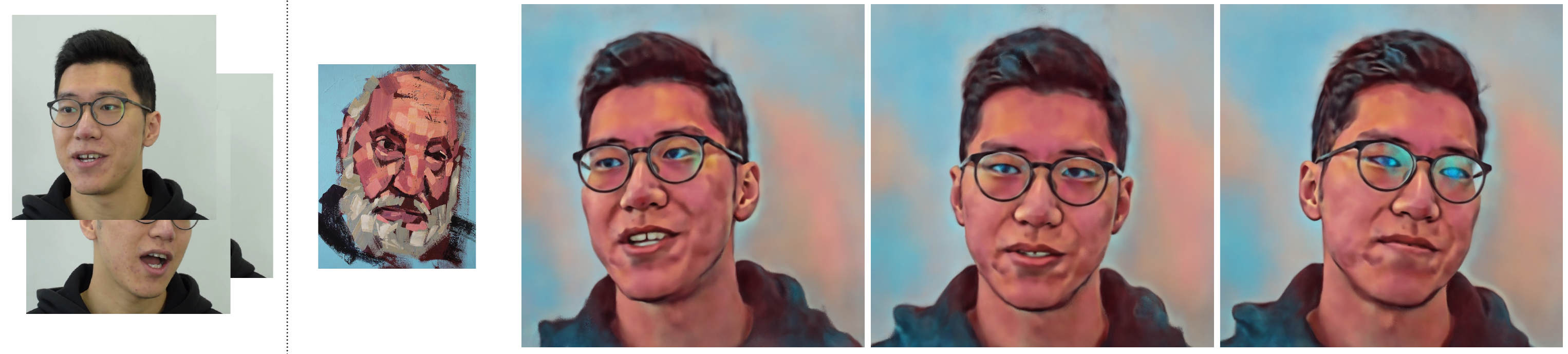}%
	\caption{\label{fig:qualitative_results_dynamic}
			SNeRF's stylization results for a dynamic avatar. Using neural implicit representations, which are compact and flexible, allows us to seamlessly extend our method to stylize this 4D dynamic avatar. Our stylized avatar can generate results that are consistent across different views and expressions.
	}
\end{figure*}
We show qualitative comparison with other approaches in Figure \ref{fig:qualitative_comparison}, in which we compare our approach with image, video and 3D-based approaches.
We encourage our readers to look at the supplementary videos to see the full effectiveness of our approach in generating cross-view consistent 3D stylization results compared to other methods.
Image style transfer approaches produce more noticeable inconsistency artifacts than the other two, since each frame is stylized independently.
Video-based approaches perform better than the image-based approaches since they take into short-term consistency. 
However, MCCNet \cite{Deng_Tang_Dong_Huang_Ma_Xu_2021} still produces noticeable artifacts when two frames are far apart, and ReReVST \cite{ReReVST2020} does not capture the reference style as well.

\new{For 3D-based approaches, we observe that StyleScene \cite{huang_2021_3d_scene_stylization}, \citet{chiang2021stylizing} and our approach generate view-consistent results since all methods aim to stylize a holistic 3D scene.
(Note that StyleScene results from the authors' model are at size 538 $\times$ 274.)
However, StyleScene's results do not capture the reference style image as well as ours, as also shown in the user study in Section \ref{sec:user_study} and Figure \ref{fig:qualitative_comparison}. 
Similarly, \citet{chiang2021stylizing}'results fail to capture the reference style well, such as the overall colour schemes or the fine-grained stippling details (Figure \ref{fig:qualitative_comparison} left).
This can be mostly explained by the fact that both of these methods stylize only scenes' \emph{appearance} instead of both geometry and appearance (see ablation study in Section \ref{ablation:freeze_geo}).
Additionally, \citet{chiang2021stylizing} only trained with small patches of size 81 $\times$ 67 out of 1008 $\times$ 550 images, which has been shown to produce results that lack global structural coherence \cite{Texler19-NPAR, huang2017adain} or diversity \cite{Wang2021}.
Finally, it is non-trivial to extend StyleScene to dynamic scenes, whereas our method can be directly applied to stylize 4D dynamic avatars (see Figure \ref{fig:qualitative_results_dynamic}).}

Figures \ref{fig:teaser}, \ref{fig:qualitative_results_DDP}, \ref{fig:qualitative_results} and \ref{fig:qualitative_results_dynamic} show additional qualitative results of our method on different scene types.
Although we choose to use the original stylization approach by \citet{neuralStyleTransfer} in this work, our scene stylization framework is not restricted to a particular stylization technique. 
For example, to stylize the dynamic avatar in Figure \ref{fig:teaser}, we use ArcaneGAN \cite{arcanegan}, which is built upon a memory-intensive StyleGAN model \cite{karras2019}.
This makes it challenging when naively combining ArcaneGAN with NeRF to perform scene stylization.
However, thanks to our alternating training approach, we can easily adopt this model to stylize a dynamic 4D avatar and produce results at high resolution (512 $\times$ 512).
\subsection{Quantitative results}
\label{sec:quantitive_results}
\subsubsection{User study}
\label{sec:user_study}
\begin{figure}[!]
	\includegraphics[width=\linewidth]{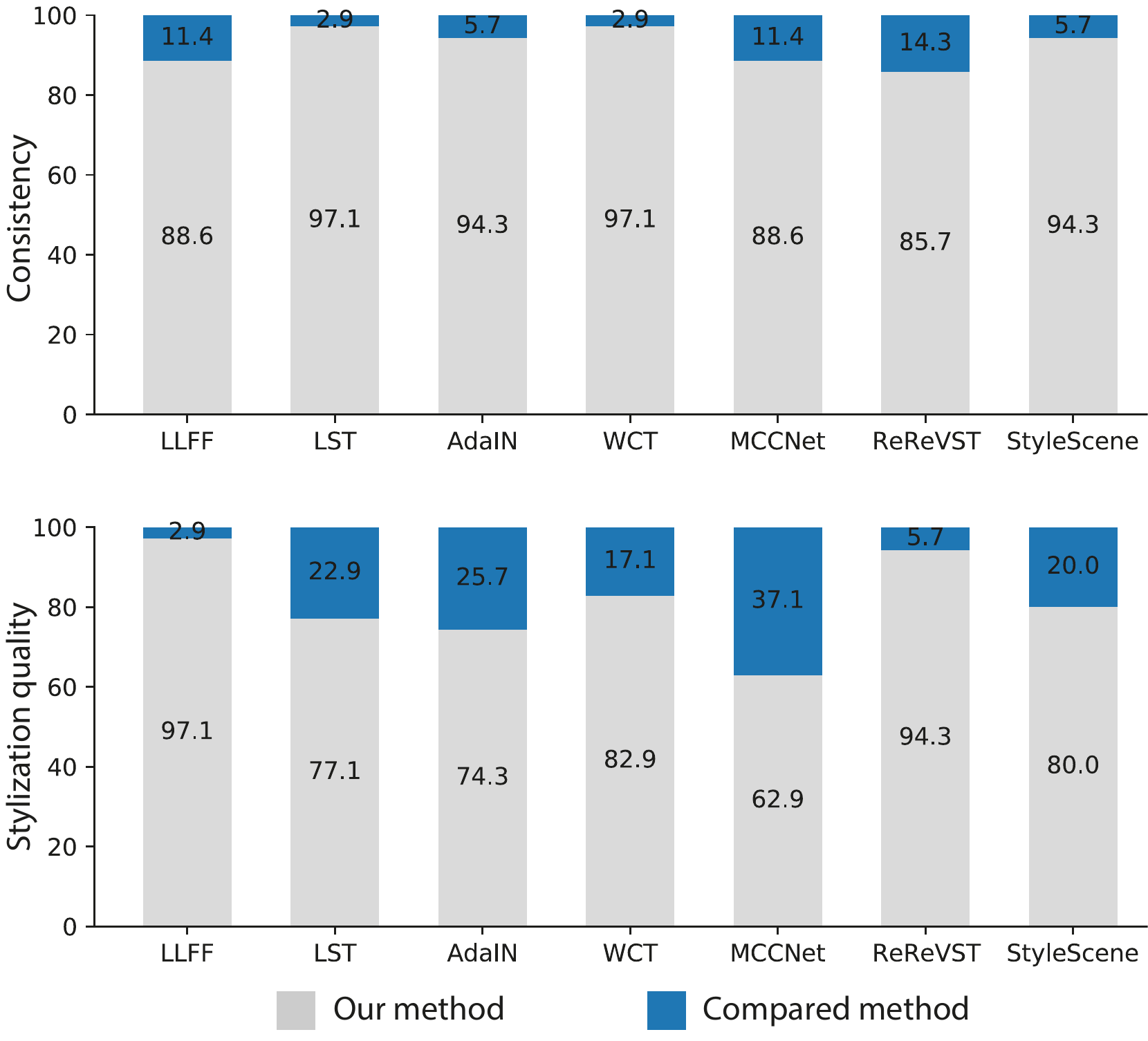}%
	\caption{\label{fig:user_consistent}
    User preference study. \new{We present two videos of novel view synthesis results, one generated by our method (grey) and one by another approach (blue - each column corresponds to one approach we are comparing against).} We ask the participant to select the one that (a) shows less flickering artifacts and (b) matches the reference style image better.
	}
\end{figure}

We conduct a user study to compare the user preference between our proposed and alternative approaches.
In particular, we want to measure users' preferences in two aspects, split into two tests: (1) which method produces more consistent results across different views (e.g., less flickering), and (2) which method matches the style of a given reference style image better.
For each question, we ask the participant to compare two videos of the same scene and style, one generated by our method and the other by one alternative method.
To generate the videos for the study, we stylize 5 scenes: \textit{Fern}, \textit{Truck}, \textit{Train}, \textit{Playground} and \textit{M60}.
We collect answers from 35 participants for both questions.
As shown in Figure \ref{fig:user_consistent}, our method (coloured in grey) performs better than other approaches on both stylization quality and multi-view consistency.
\subsubsection{Cross-view consistency}
One of the main advantages of using implicit scene representation for stylization is cross-view consistency.
To test the performance of our approach, we adopt a similar strategy to \citet{Lai-ECCV-2018} to measure the consistency between different novel views of the stylized 3D scene.
In particular, we create testing videos where each frame is a rendered image at a novel view of our stylized scene.
We then compute the optical flow  and the occlusion mask $O$ using two ground truth views $I^{real}_{i}$ and $I^{real}_{i + \delta}$ rendered from a pre-trained NeRF scene. 
Note that unlike \citet{Lai-ECCV-2018} who use FlowNet2 \cite{flownet}, we use RAFT \cite{RAFT}, a state-of-the-art method to predict optical flow between two views.
Secondly, we warp a stylized view $I^s_{i}$ to obtain a new view $\hat{I}^s_{i + \delta}$ using the optical flow.
Finally, we compute the error between the novel view obtained from our stylized scene $I^s_{i + \delta}$, and the previously computed $\hat{I}^s_{i + \delta}$ using:
\begin{equation}
 E_{consistency} (I^s_{i + \delta}, \hat{I}^s_{i + \delta}) = \frac{1}{|O'|}\norm{ \dot I^s_{i + \delta} - \hat{I}^s_{i + \delta}}_2^2 
\end{equation}
where $|O'|$ denotes the number of non-occluded pixels calculated using $O$.
Following \citet{huang_2021_3d_scene_stylization} and \citet{chiang2021stylizing}, we measure both \textit{short-range} and \textit{long-range} consistency between different testing video frames.
Specifically, we compute the error between $i^{th}$ and $(i + 1)^{th}$ video frames to measure \textit{short-range} consistency, and between $i^{th}$ and $(i + 7)^{th}$ frames for \textit{long-range} consistency.

We show our results for short and long-range consistency measurement in Table \ref{tab:shortrange} and \ref{tab:longrange} respectively. 
We compare our method with image-based approaches (AdaIN, WCT and LST), as well as video-based approaches (ReReVST and MCCNet), and report the average errors of 12 diverse styles.
Unfortunately, we could not match the camera path and resolution for the results of StyleScene \cite{huang_2021_3d_scene_stylization}, and thus do not include this work in this comparison.
In general, the image stylization alternative methods produce worse results than video-based and 3D-based methods (ours).
We observe that ReReVST produces competitive results with our method.
However, as shown in Figure \ref{fig:qualitative_comparison}, and user study in Section \ref{sec:user_study}, ReReVST does not capture the reference style well.
\begin{table}[h]%
\caption{Qualitative comparisons on short-range consistency. We compute the consistency score (the lower the better) between two nearby stylized novel views. The best result is in bold and the second best is underscored.}
\label{tab:shortrange}
\begin{minipage}{\columnwidth}
\begin{center}
\begin{tabular}{*{6}{c}}
  \toprule
  Methods & Truck & Playground & M60 & Train\\ \midrule
  AdaIN & 0.043  & 0.044 & 0.054 & 0.039\\
  WCT & 0.064 & 0.063 & 0.084 & 0.056\\
  LST& 0.027 & 0.026 & 0.032 & 0.024\\
  \midrule
  ReReVST& \underline{0.010} & \underline{0.009} & \textbf{0.010} & \underline{0.015}\\
  MCCNet& 0.025 & 0.025 & 0.028 & 0.021\\
  \midrule
  \name (Ours) & \textbf{0.009} & \textbf{0.004} & \underline{0.012} & \textbf{0.008}\\
  \bottomrule
\end{tabular}
\end{center}
\bigskip\centering
\end{minipage}
\end{table}%
\begin{table}[h]%
\caption{Qualitative comparisons on long-range consistency. We compute the consistency score (the lower the better) between two far-away stylized novel views. The best result is in bold and the second best is underscored.}
\label{tab:longrange}
\begin{minipage}{\columnwidth}
\begin{center}
\begin{tabular}{*{6}{c}}
  \toprule
  Methods & Truck & Playground & M60 & Train\\ \midrule
  AdaIN& 0.059  & 0.060 & 0.075 & 0.062\\
  WCT& 0.084 & 0.087 & 0.110 & 0.082\\
  LST& 0.037 & 0.032 & 0.042 & 0.036\\
  \midrule
  ReReVST&       \textbf{0.015} & \underline{0.015} & \textbf{0.016} & \underline{0.024}\\
  MCCNet& 0.035 & 0.030 & 0.039 & 0.034\\
  \midrule
  \name (Ours)& \underline{0.026} & \textbf{0.010} & \underline{0.032} & \textbf{0.016}\\
  \bottomrule
\end{tabular}
\end{center}
\bigskip\centering
\end{minipage}
\end{table}%
As shown in the user studies and consistency measurements, \name can stylize 3D scenes to generate novel views faithful to both the reference style and the original scene content while maintaining cross-view consistency.
\subsubsection{Freezing geometry}
\label{ablation:freeze_geo}
\begin{figure*}[h!]
	\includegraphics[width=1.0\linewidth]{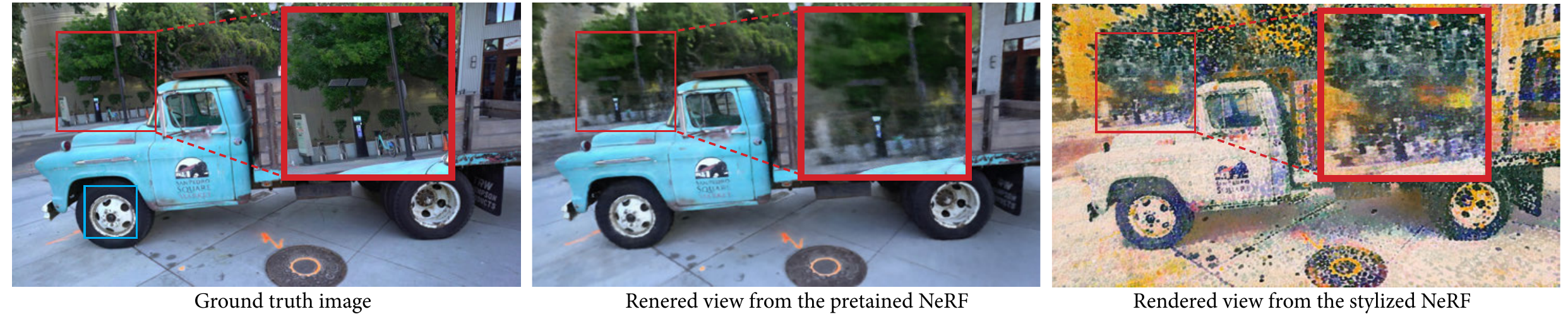}
	\caption{\label{fig:limitation} The quality of our stylized results partly depends on the quality of the underlying implicit scene function. If the scene function fails to capture sharp details (shown in the red box), the stylized results will be blurry.}
\end{figure*}
\begin{figure}
	\includegraphics[width=\linewidth]{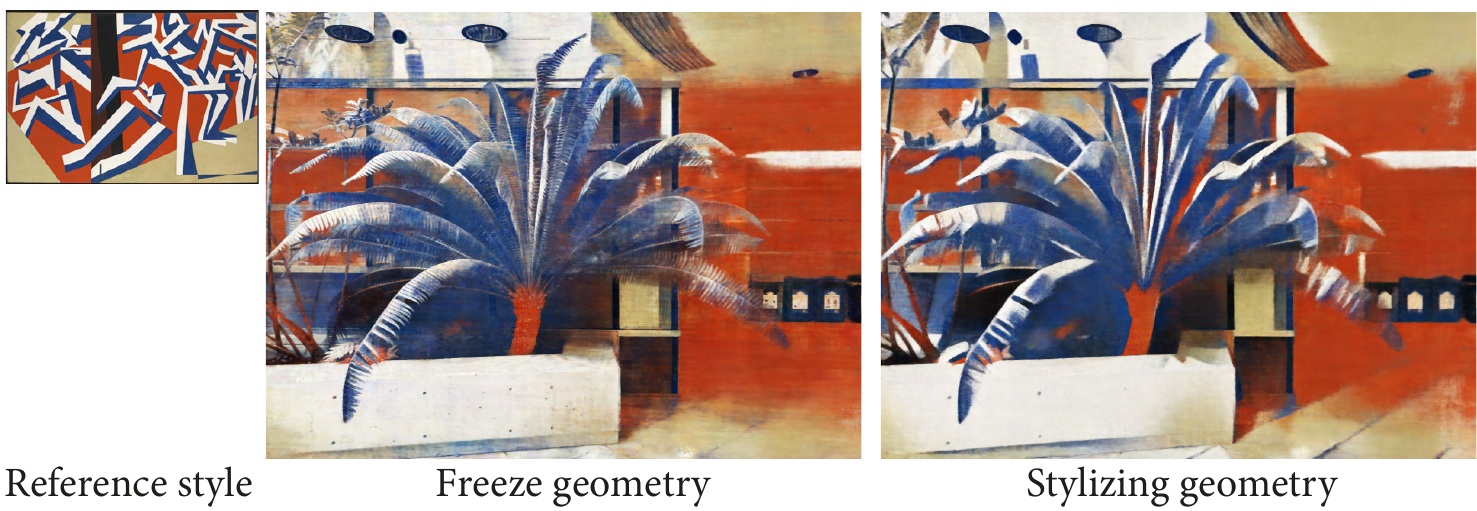}%
	\caption{\label{fig:ablation_freezeGeometry} The effects of only updating the appearance of NeRF. Stylizing both geometry and appearance leads to results with sharper details and closer to the reference style, especially for more abstract reference style images.}
\end{figure}
Recent 3D scene stylization approaches focus only on stylizing the appearance instead of both the underlying geometry and appearance~\cite{huang_2021_3d_scene_stylization, chiang2021stylizing}.
However, \citet{9577906} and \citet{Kim20DST} show that geometry is also a component of style, and thus stylizing both style and geometry leads to stylized results that are closer to the target style.
Therefore, our method stylizes both the appearance and geometry (represented as density) of the scene functions. 
Figure \ref{fig:ablation_freezeGeometry} shows that this produces stylized results that are closer to the target style, especially when the style is more abstract or contains lots of fine-grained details.
Meanwhile, when we only stylize the appearance \new{(by keeping the weights of NeRF's shared and opacity branch fixed, and only updating the weights of the RGB branch)}, the results only capture the style's color scheme.
\subsubsection{Alternating training scheme.}
\label{ablation:trainNeRF_oneStep}
In addition to addressing the memory limitations, our alternating training framework can also produce stylization results that capture the reference style better.
\old{The 3D scene stylization process comprises two main steps. The first step is image based stylization, which allows us to perform stylization using a reference image.
However, this does not guarantee multi-view consistency since each view is processed independently.
The second step is scene stylization, which modulates the scene such that its rendered images match the set of stylized images.}
The 3D scene stylization process comprises two main steps: image stylization, which allows us to perform stylization using a reference image but does not guarantee multi-view consistency, and
scene stylization, which modulates the scene to match the set of stylized views and maintain consistency.
In our method, we alternate between these two steps for a few iterations, where one iteration comprises stylization and NeRF training.

We show that naively training a scene function using a set of inconsistent stylized images leads to cross-view consistent results, but fails to capture the target style, similar to the novel view synthesis results by LLFF.
This is the equivalent of performing only one iteration of our approach. 
\begin{figure}[h!]
	\includegraphics[width=\linewidth]{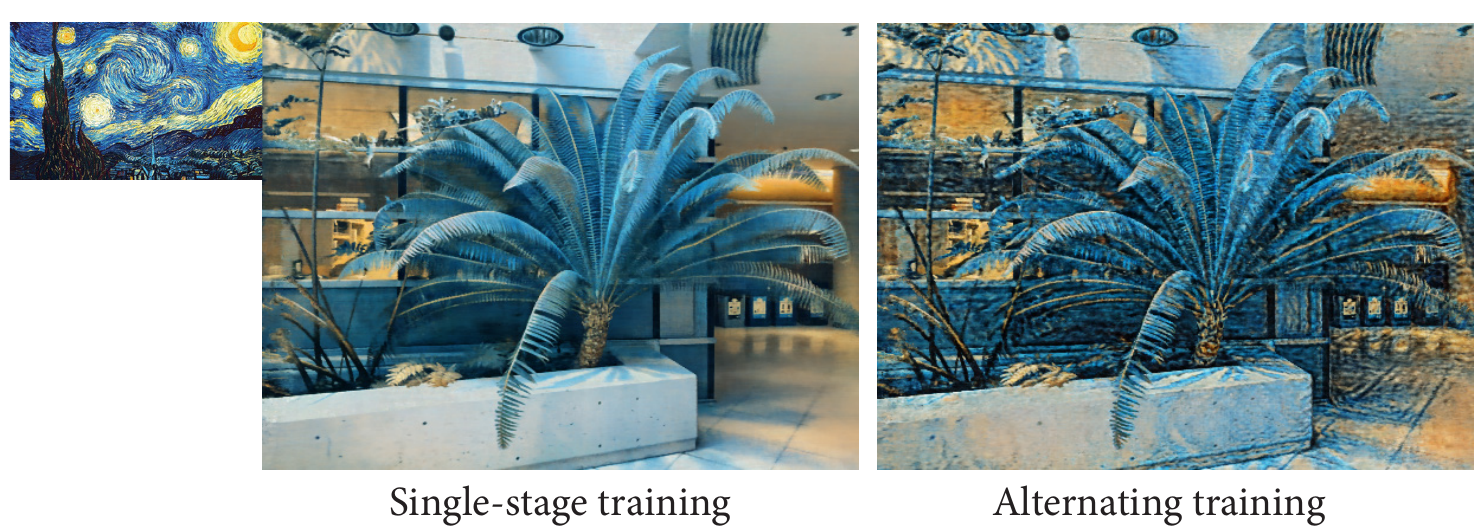}%
	\caption{\label{fig:ablation_alternating} Alternating between training NeRF and stylization leads to better results than a single-stage training using individually stylized images.}
\end{figure}
\old{However, as shown in Figure \ref{fig:ablation_alternating}, if we then render the now more stylized scene function, and repeat the same process of stylization and NeRF training for a few more iterations, we get better results that achieve both multi-view consistency and fine-grained details of the reference style.}
However, as shown in Figure \ref{fig:ablation_alternating}, if we repeat the same process of stylization and NeRF training for more iterations, we get better results that are multi-view consistent and capture fine-grained details of the reference style.

\section{Discussion}
In this paper, we have shown both qualitatively and quantitatively the advantages of \name in terms of generating multi-view consistent results, compared to image and video style transfer methods~\cite{huang2017adain, ReReVST2020, WCT-NIPS-2017, li2018learning, Deng_Tang_Dong_Huang_Ma_Xu_2021}.
We also show that our method generates better stylization results than other 3D-based approaches ~\cite{huang_2021_3d_scene_stylization, chiang2021stylizing}, which can be mostly attributed to our method's flexibility in stylizing both the 3D scene geometry and appearance. 
Unlike these existing methods that focus on static scenes, our method can also stylize dynamic content, such as 4D avatars.

We observe that the quality of our stylized results partly depends on the quality of the scene function trained with RGB images, as also observed by \citet{chiang2021stylizing}. For example, in Figure \ref{fig:limitation}, the underlying NeRF model fails to capture sharp details of the vegetation in the background, compared to the ground truth RGB image. This eventually leads to blurry stylized results. 

Our method adopts NeRF which is more time-consuming and computationally demanding than point clouds. \new{Depending on the resolution of the scene, on a single Nvidia V100 GPU, training (including training the original NeRF models) can take 3-5 days and rendering an image at size 1008 $\times$ 548 can take 55 seconds.} We believe that our alternating stylization framework will enable quick adoption of fast emerging advances in research on NeRF to improve quality \cite{mipnerf360, barron2021mipnerf} and speed \cite{hedman2021snerg, neff2021donerf, mueller2022instant}, as well as style transfer methods to improve the fidelity and variety of stylization results~\cite{gal2021stylegannada, Texler19-NPAR}. Note that, even though it takes a while to train a stylized NeRF using our method, once trained, they can be ``baked''~\cite{hedman2021baking} for real-time rendering in AR, VR and MR applications.

Apart from RGB images, future work can explore using additional segmentation or depth maps (queried from NeRF models) for stylization~\cite{liu2017, wang2020glstylenet}.
Secondly, we currently stylize each scene independently, and cannot apply an arbitrary style to each scene without restarting the optimization process. 
Therefore, it will be an interesting direction to combine our framework with recent work on arbitrary style transfer.

\section{Conclusion}
In this work, we present a method for 3D scene stylization using implicit neural representations (NeRF).
This provides a strong inductive bias to produce stylized multi-view consistent results that also match a target style image well.
Additionally, our alternating stylization method enables us to make full use of our hardware memory capability to stylize both static and dynamic 3D scenes, allowing us to both generate images at higher resolution and adopt more expressive image style transfer methods.
As NeRF increasingly attracts more research in improving generalisation, quality, and speed in both training and test time, we believe that using implicit scene representations for 3D scene stylization will open up a wide range of exciting applications for AR, VR and MR.

\bibliographystyle{ACM-Reference-Format}
\bibliography{ref}


\begin{thebibliography}{78}


\ifx \showCODEN    \undefined \def \showCODEN     #1{\unskip}     \fi
\ifx \showDOI      \undefined \def \showDOI       #1{#1}\fi
\ifx \showISBNx    \undefined \def \showISBNx     #1{\unskip}     \fi
\ifx \showISBNxiii \undefined \def \showISBNxiii  #1{\unskip}     \fi
\ifx \showISSN     \undefined \def \showISSN      #1{\unskip}     \fi
\ifx \showLCCN     \undefined \def \showLCCN      #1{\unskip}     \fi
\ifx \shownote     \undefined \def \shownote      #1{#1}          \fi
\ifx \showarticletitle \undefined \def \showarticletitle #1{#1}   \fi
\ifx \showURL      \undefined \def \showURL       {\relax}        \fi
\providecommand\bibfield[2]{#2}
\providecommand\bibinfo[2]{#2}
\providecommand\natexlab[1]{#1}
\providecommand\showeprint[2][]{arXiv:#2}

\bibitem[\protect\citeauthoryear{Aliev, Sevastopolsky, Kolos, Ulyanov, and
  Lempitsky}{Aliev et~al\mbox{.}}{2020}]%
        {aliev2020neural}
\bibfield{author}{\bibinfo{person}{Kara-Ali Aliev}, \bibinfo{person}{Artem
  Sevastopolsky}, \bibinfo{person}{Maria Kolos}, \bibinfo{person}{Dmitry
  Ulyanov}, {and} \bibinfo{person}{Victor Lempitsky}.}
  \bibinfo{year}{2020}\natexlab{}.
\newblock \showarticletitle{Neural point-based graphics}. In
  \bibinfo{booktitle}{\emph{Proceedings of the European Conference on Computer
  Vision}}. \bibinfo{pages}{696--712}.
\newblock


\bibitem[\protect\citeauthoryear{Barron, Mildenhall, Tancik, Hedman,
  Martin-Brualla, and Srinivasan}{Barron et~al\mbox{.}}{2021}]%
        {barron2021mipnerf}
\bibfield{author}{\bibinfo{person}{Jonathan~T. Barron}, \bibinfo{person}{Ben
  Mildenhall}, \bibinfo{person}{Matthew Tancik}, \bibinfo{person}{Peter
  Hedman}, \bibinfo{person}{Ricardo Martin-Brualla}, {and}
  \bibinfo{person}{Pratul~P. Srinivasan}.} \bibinfo{year}{2021}\natexlab{}.
\newblock \showarticletitle{Mip-NeRF: A Multiscale Representation for
  Anti-Aliasing Neural Radiance Fields}. In
  \bibinfo{booktitle}{\emph{Proceedings of the IEEE/CVF International
  Conference on Computer Vision (ICCV)}}. \bibinfo{pages}{5855--5864}.
\newblock


\bibitem[\protect\citeauthoryear{Barron, Mildenhall, Verbin, Srinivasan, and
  Hedman}{Barron et~al\mbox{.}}{2022}]%
        {mipnerf360}
\bibfield{author}{\bibinfo{person}{Jonathan~T. Barron}, \bibinfo{person}{Ben
  Mildenhall}, \bibinfo{person}{Dor Verbin}, \bibinfo{person}{Pratul~P.
  Srinivasan}, {and} \bibinfo{person}{Peter Hedman}.}
  \bibinfo{year}{2022}\natexlab{}.
\newblock \showarticletitle{Mip-NeRF 360: Unbounded Anti-Aliased Neural
  Radiance Fields}.
\newblock \bibinfo{journal}{\emph{Proceedings of the IEEE/CVF Conference on
  Computer Vision and Pattern Recognition}} (\bibinfo{year}{2022}).
\newblock


\bibitem[\protect\citeauthoryear{Broxton, Flynn, Overbeck, Erickson, Hedman,
  Duvall, Dourgarian, Busch, Whalen, and Debevec}{Broxton
  et~al\mbox{.}}{2020}]%
        {Broxton2020}
\bibfield{author}{\bibinfo{person}{Michael Broxton}, \bibinfo{person}{John
  Flynn}, \bibinfo{person}{Ryan Overbeck}, \bibinfo{person}{Daniel Erickson},
  \bibinfo{person}{Peter Hedman}, \bibinfo{person}{Matthew Duvall},
  \bibinfo{person}{Jason Dourgarian}, \bibinfo{person}{Jay Busch},
  \bibinfo{person}{Matt Whalen}, {and} \bibinfo{person}{Paul Debevec}.}
  \bibinfo{year}{2020}\natexlab{}.
\newblock \showarticletitle{Immersive Light Field Video with a Layered Mesh
  Representation}.
\newblock \bibinfo{journal}{\emph{ACM Trans. Graph.}} \bibinfo{volume}{39},
  \bibinfo{number}{4} (\bibinfo{date}{July} \bibinfo{year}{2020}),
  \bibinfo{pages}{86:1--15}.
\newblock


\bibitem[\protect\citeauthoryear{Buehler, Bosse, McMillan, Gortler, and
  Cohen}{Buehler et~al\mbox{.}}{2001}]%
        {buehler2001unstructured}
\bibfield{author}{\bibinfo{person}{Chris Buehler}, \bibinfo{person}{Michael
  Bosse}, \bibinfo{person}{Leonard McMillan}, \bibinfo{person}{Steven Gortler},
  {and} \bibinfo{person}{Michael Cohen}.} \bibinfo{year}{2001}\natexlab{}.
\newblock \showarticletitle{Unstructured lumigraph rendering}. In
  \bibinfo{booktitle}{\emph{Proceedings of the 28th annual conference on
  Computer graphics and interactive techniques}}. \bibinfo{pages}{425--432}.
\newblock


\bibitem[\protect\citeauthoryear{Cao, Wang, Nagao, and Nakamura}{Cao
  et~al\mbox{.}}{2020}]%
        {Cao_2020_WACV}
\bibfield{author}{\bibinfo{person}{Xu Cao}, \bibinfo{person}{Weimin Wang},
  \bibinfo{person}{Katashi Nagao}, {and} \bibinfo{person}{Ryosuke Nakamura}.}
  \bibinfo{year}{2020}\natexlab{}.
\newblock \showarticletitle{PSNet: A Style Transfer Network for Point Cloud
  Stylization on Geometry and Color}. In \bibinfo{booktitle}{\emph{The IEEE
  Winter Conference on Applications of Computer Vision}}.
\newblock


\bibitem[\protect\citeauthoryear{Chaurasia, Duchene, Sorkine-Hornung, and
  Drettakis}{Chaurasia et~al\mbox{.}}{2013}]%
        {Chaurasia:13}
\bibfield{author}{\bibinfo{person}{Gaurav Chaurasia}, \bibinfo{person}{Sylvain
  Duchene}, \bibinfo{person}{Olga Sorkine-Hornung}, {and}
  \bibinfo{person}{George Drettakis}.} \bibinfo{year}{2013}\natexlab{}.
\newblock \showarticletitle{Depth Synthesis and Local Warps for Plausible
  Image-Based Navigation}.
\newblock \bibinfo{journal}{\emph{ACM Trans. Graph.}} \bibinfo{volume}{32},
  \bibinfo{number}{3}, Article \bibinfo{articleno}{30} (\bibinfo{date}{jul}
  \bibinfo{year}{2013}).
\newblock


\bibitem[\protect\citeauthoryear{Chen, Liao, Yuan, Yu, and Hua}{Chen
  et~al\mbox{.}}{2017}]%
        {8237388}
\bibfield{author}{\bibinfo{person}{Dongdong Chen}, \bibinfo{person}{Jing Liao},
  \bibinfo{person}{Lu Yuan}, \bibinfo{person}{Nenghai Yu}, {and}
  \bibinfo{person}{Gang Hua}.} \bibinfo{year}{2017}\natexlab{}.
\newblock \showarticletitle{Coherent Online Video Style Transfer}. In
  \bibinfo{booktitle}{\emph{Proceedings of the IEEE/CVF International
  Conference on Computer Vision}}. \bibinfo{pages}{1114--1123}.
\newblock


\bibitem[\protect\citeauthoryear{Chen, Yuan, Liao, Yu, and Hua}{Chen
  et~al\mbox{.}}{2018}]%
        {Chen_2018_CVPR}
\bibfield{author}{\bibinfo{person}{Dongdong Chen}, \bibinfo{person}{Lu Yuan},
  \bibinfo{person}{Jing Liao}, \bibinfo{person}{Nenghai Yu}, {and}
  \bibinfo{person}{Gang Hua}.} \bibinfo{year}{2018}\natexlab{}.
\newblock \showarticletitle{Stereoscopic Neural Style Transfer}. In
  \bibinfo{booktitle}{\emph{Proceedings of the IEEE/CVF Conference on Computer
  Vision and Pattern Recognition}}.
\newblock


\bibitem[\protect\citeauthoryear{Chen, Zhang, Wang, Shu, Xu, and Xu}{Chen
  et~al\mbox{.}}{2020}]%
        {opticalFlowdistillation}
\bibfield{author}{\bibinfo{person}{Xinghao Chen}, \bibinfo{person}{Yiman
  Zhang}, \bibinfo{person}{Yunhe Wang}, \bibinfo{person}{Han Shu},
  \bibinfo{person}{Chunjing Xu}, {and} \bibinfo{person}{Chang Xu}.}
  \bibinfo{year}{2020}\natexlab{}.
\newblock \showarticletitle{Optical Flow Distillation: Towards Efficient and
  Stable Video Style Transfer}. In \bibinfo{booktitle}{\emph{Proceedings of the
  European Conference on Computer Vision}} \emph{(\bibinfo{series}{Lecture
  Notes in Computer Science}, Vol.~\bibinfo{volume}{12351})},
  \bibfield{editor}{\bibinfo{person}{Andrea Vedaldi}, \bibinfo{person}{Horst
  Bischof}, \bibinfo{person}{Thomas Brox}, {and} \bibinfo{person}{Jan{-}Michael
  Frahm}} (Eds.). \bibinfo{publisher}{Springer}, \bibinfo{pages}{614--630}.
\newblock


\bibitem[\protect\citeauthoryear{Chiang, Tsai, Tseng, Lai, and Chiu}{Chiang
  et~al\mbox{.}}{2022}]%
        {chiang2021stylizing}
\bibfield{author}{\bibinfo{person}{Pei-Ze Chiang}, \bibinfo{person}{Meng-Shiun
  Tsai}, \bibinfo{person}{Hung-Yu Tseng}, \bibinfo{person}{Wei-Sheng Lai},
  {and} \bibinfo{person}{Wei-Chen Chiu}.} \bibinfo{year}{2022}\natexlab{}.
\newblock \showarticletitle{Stylizing 3D Scene via Implicit Representation and
  HyperNetwork}.
\newblock  (\bibinfo{date}{January} \bibinfo{year}{2022}),
  \bibinfo{pages}{1475--1484}.
\newblock


\bibitem[\protect\citeauthoryear{Dellaert and Yen-Chen}{Dellaert and
  Yen-Chen}{2021}]%
        {dellaert2021neural}
\bibfield{author}{\bibinfo{person}{Frank Dellaert} {and} \bibinfo{person}{Lin
  Yen-Chen}.} \bibinfo{year}{2021}\natexlab{}.
\newblock \bibinfo{title}{Neural Volume Rendering: NeRF And Beyond}.
\newblock
\newblock
\showeprint[arxiv]{2101.05204}~[cs.CV]


\bibitem[\protect\citeauthoryear{Deng, Tang, Dong, Huang, Ma, and Xu}{Deng
  et~al\mbox{.}}{2021}]%
        {Deng_Tang_Dong_Huang_Ma_Xu_2021}
\bibfield{author}{\bibinfo{person}{Yingying Deng}, \bibinfo{person}{Fan Tang},
  \bibinfo{person}{Weiming Dong}, \bibinfo{person}{Haibin Huang},
  \bibinfo{person}{Chongyang Ma}, {and} \bibinfo{person}{Changsheng Xu}.}
  \bibinfo{year}{2021}\natexlab{}.
\newblock \showarticletitle{Arbitrary Video Style Transfer via Multi-Channel
  Correlation}.
\newblock \bibinfo{journal}{\emph{Proceedings of the AAAI Conference on
  Artificial Intelligence}} \bibinfo{volume}{35}, \bibinfo{number}{2}
  (\bibinfo{date}{May} \bibinfo{year}{2021}), \bibinfo{pages}{1210--1217}.
\newblock


\bibitem[\protect\citeauthoryear{Fi\v{s}er, Jamri\v{s}ka, Luk\'{a}\v{c},
  Shechtman, Asente, Lu, and S\'{y}kora}{Fi\v{s}er et~al\mbox{.}}{2016}]%
        {Stylit}
\bibfield{author}{\bibinfo{person}{Jakub Fi\v{s}er},
  \bibinfo{person}{Ond\v{r}ej Jamri\v{s}ka}, \bibinfo{person}{Michal
  Luk\'{a}\v{c}}, \bibinfo{person}{Eli Shechtman}, \bibinfo{person}{Paul
  Asente}, \bibinfo{person}{Jingwan Lu}, {and} \bibinfo{person}{Daniel
  S\'{y}kora}.} \bibinfo{year}{2016}\natexlab{}.
\newblock \showarticletitle{StyLit: Illumination-Guided Example-Based
  Stylization of 3D Renderings}.
\newblock \bibinfo{journal}{\emph{ACM Trans. Graph.}} \bibinfo{volume}{35},
  \bibinfo{number}{4}, Article \bibinfo{articleno}{92} (\bibinfo{date}{jul}
  \bibinfo{year}{2016}), \bibinfo{numpages}{11}~pages.
\newblock
\showISSN{0730-0301}


\bibitem[\protect\citeauthoryear{Flynn, Neulander, Philbin, and Snavely}{Flynn
  et~al\mbox{.}}{2016}]%
        {flynn2016deepstereo}
\bibfield{author}{\bibinfo{person}{John Flynn}, \bibinfo{person}{Ivan
  Neulander}, \bibinfo{person}{James Philbin}, {and} \bibinfo{person}{Noah
  Snavely}.} \bibinfo{year}{2016}\natexlab{}.
\newblock \showarticletitle{Deepstereo: Learning to predict new views from the
  world's imagery}. In \bibinfo{booktitle}{\emph{Proceedings of the IEEE/CVF
  Conference on Computer Vision and Pattern Recognition}}.
  \bibinfo{pages}{5515--5524}.
\newblock


\bibitem[\protect\citeauthoryear{Gafni, Thies, Zollh{\"o}fer, and
  Nie{\ss}ner}{Gafni et~al\mbox{.}}{2021}]%
        {Gafni_2021_CVPR}
\bibfield{author}{\bibinfo{person}{Guy Gafni}, \bibinfo{person}{Justus Thies},
  \bibinfo{person}{Michael Zollh{\"o}fer}, {and} \bibinfo{person}{Matthias
  Nie{\ss}ner}.} \bibinfo{year}{2021}\natexlab{}.
\newblock \showarticletitle{Dynamic Neural Radiance Fields for Monocular 4D
  Facial Avatar Reconstruction}. In \bibinfo{booktitle}{\emph{Proceedings of
  the IEEE/CVF Conference on Computer Vision and Pattern Recognition}}.
  \bibinfo{pages}{8649--8658}.
\newblock


\bibitem[\protect\citeauthoryear{Gal, Patashnik, Maron, Chechik, and
  Cohen-Or}{Gal et~al\mbox{.}}{2021}]%
        {gal2021stylegannada}
\bibfield{author}{\bibinfo{person}{Rinon Gal}, \bibinfo{person}{Or Patashnik},
  \bibinfo{person}{Haggai Maron}, \bibinfo{person}{Gal Chechik}, {and}
  \bibinfo{person}{Daniel Cohen-Or}.} \bibinfo{year}{2021}\natexlab{}.
\newblock \bibinfo{title}{StyleGAN-NADA: CLIP-Guided Domain Adaptation of Image
  Generators}.
\newblock
\newblock
\showeprint[arxiv]{2108.00946}~[cs.CV]


\bibitem[\protect\citeauthoryear{Gao, Gu, Zhang, and Yu}{Gao
  et~al\mbox{.}}{2018}]%
        {GaoACCV}
\bibfield{author}{\bibinfo{person}{Chang Gao}, \bibinfo{person}{Derun Gu},
  \bibinfo{person}{Fangjun Zhang}, {and} \bibinfo{person}{Yizhou Yu}.}
  \bibinfo{year}{2018}\natexlab{}.
\newblock \showarticletitle{ReCoNet: Real-time Coherent Video Style Transfer
  Network}. In \bibinfo{booktitle}{\emph{Proceedings of the 18th Asian
  Conference on Computer Vision}}, Vol.~\bibinfo{volume}{abs/1807.01197}.
\newblock


\bibitem[\protect\citeauthoryear{Gatys, Ecker, and Bethge}{Gatys
  et~al\mbox{.}}{2016}]%
        {neuralStyleTransfer}
\bibfield{author}{\bibinfo{person}{Leon~A. Gatys},
  \bibinfo{person}{Alexander~S. Ecker}, {and} \bibinfo{person}{Matthias
  Bethge}.} \bibinfo{year}{2016}\natexlab{}.
\newblock \showarticletitle{Image Style Transfer Using Convolutional Neural
  Networks}. In \bibinfo{booktitle}{\emph{Proceedings of the IEEE/CVF
  Conference on Computer Vision and Pattern Recognition}}.
  \bibinfo{pages}{2414--2423}.
\newblock


\bibitem[\protect\citeauthoryear{Gong, Huang, Ma, Shen, Liu, and Zhang}{Gong
  et~al\mbox{.}}{2018}]%
        {Gong_2018_ECCV}
\bibfield{author}{\bibinfo{person}{Xinyu Gong}, \bibinfo{person}{Haozhi Huang},
  \bibinfo{person}{Lin Ma}, \bibinfo{person}{Fumin Shen}, \bibinfo{person}{Wei
  Liu}, {and} \bibinfo{person}{Tong Zhang}.} \bibinfo{year}{2018}\natexlab{}.
\newblock \showarticletitle{Neural Stereoscopic Image Style Transfer}. In
  \bibinfo{booktitle}{\emph{Proceedings of the European Conference on Computer
  Vision}}.
\newblock


\bibitem[\protect\citeauthoryear{Gortler, Grzeszczuk, Szeliski, and
  Cohen}{Gortler et~al\mbox{.}}{1996}]%
        {Gortler:96}
\bibfield{author}{\bibinfo{person}{Steven~J. Gortler}, \bibinfo{person}{Radek
  Grzeszczuk}, \bibinfo{person}{Richard Szeliski}, {and}
  \bibinfo{person}{Michael~F. Cohen}.} \bibinfo{year}{1996}\natexlab{}.
\newblock \showarticletitle{The Lumigraph}. In
  \bibinfo{booktitle}{\emph{Proceedings of the 23rd Annual Conference on
  Computer Graphics and Interactive Techniques}}. \bibinfo{pages}{43–54}.
\newblock


\bibitem[\protect\citeauthoryear{Hauptfleisch, Texler, Texler,
  K\v{r}iv\'{a}nek, and S\'{y}kora}{Hauptfleisch et~al\mbox{.}}{2020}]%
        {Hauptfleisch20-PG}
\bibfield{author}{\bibinfo{person}{Filip Hauptfleisch},
  \bibinfo{person}{Ond\v{r}ej Texler}, \bibinfo{person}{Aneta Texler},
  \bibinfo{person}{Jaroslav K\v{r}iv\'{a}nek}, {and} \bibinfo{person}{Daniel
  S\'{y}kora}.} \bibinfo{year}{2020}\natexlab{}.
\newblock \showarticletitle{{StyleProp}: {Real}-time Example-based Stylization
  of 3D Models}.
\newblock \bibinfo{journal}{\emph{Computer Graphics Forum}}
  \bibinfo{volume}{39}, \bibinfo{number}{7} (\bibinfo{year}{2020}),
  \bibinfo{pages}{575--586}.
\newblock


\bibitem[\protect\citeauthoryear{Hedman, Philip, Price, Frahm, Drettakis, and
  Brostow}{Hedman et~al\mbox{.}}{2018}]%
        {Hedman18}
\bibfield{author}{\bibinfo{person}{Peter Hedman}, \bibinfo{person}{Julien
  Philip}, \bibinfo{person}{True Price}, \bibinfo{person}{Jan-Michael Frahm},
  \bibinfo{person}{George Drettakis}, {and} \bibinfo{person}{Gabriel Brostow}.}
  \bibinfo{year}{2018}\natexlab{}.
\newblock \showarticletitle{Deep Blending for Free-Viewpoint Image-Based
  Rendering}.
\newblock \bibinfo{journal}{\emph{ACM Trans. Graph.}} \bibinfo{volume}{37},
  \bibinfo{number}{6}, Article \bibinfo{articleno}{257} (\bibinfo{date}{dec}
  \bibinfo{year}{2018}).
\newblock


\bibitem[\protect\citeauthoryear{Hedman, Srinivasan, Mildenhall, Barron, and
  Debevec}{Hedman et~al\mbox{.}}{2021a}]%
        {hedman2021snerg}
\bibfield{author}{\bibinfo{person}{Peter Hedman}, \bibinfo{person}{Pratul~P.
  Srinivasan}, \bibinfo{person}{Ben Mildenhall}, \bibinfo{person}{Jonathan~T.
  Barron}, {and} \bibinfo{person}{Paul Debevec}.}
  \bibinfo{year}{2021}\natexlab{a}.
\newblock \showarticletitle{Baking Neural Radiance Fields for Real-Time View
  Synthesis}.
\newblock  (\bibinfo{date}{October} \bibinfo{year}{2021}),
  \bibinfo{pages}{5875--5884}.
\newblock


\bibitem[\protect\citeauthoryear{Hedman, Srinivasan, Mildenhall, Barron, and
  Debevec}{Hedman et~al\mbox{.}}{2021b}]%
        {hedman2021baking}
\bibfield{author}{\bibinfo{person}{Peter Hedman}, \bibinfo{person}{Pratul~P.
  Srinivasan}, \bibinfo{person}{Ben Mildenhall}, \bibinfo{person}{Jonathan~T.
  Barron}, {and} \bibinfo{person}{Paul Debevec}.}
  \bibinfo{year}{2021}\natexlab{b}.
\newblock \showarticletitle{Baking Neural Radiance Fields for Real-Time View
  Synthesis}.
\newblock  (\bibinfo{date}{October} \bibinfo{year}{2021}),
  \bibinfo{pages}{5875--5884}.
\newblock


\bibitem[\protect\citeauthoryear{Hertzmann, Jacobs, Oliver, Curless, and
  Salesin}{Hertzmann et~al\mbox{.}}{2001}]%
        {hertzmann2001image}
\bibfield{author}{\bibinfo{person}{Aaron Hertzmann}, \bibinfo{person}{Charles~E
  Jacobs}, \bibinfo{person}{Nuria Oliver}, \bibinfo{person}{Brian Curless},
  {and} \bibinfo{person}{David~H Salesin}.} \bibinfo{year}{2001}\natexlab{}.
\newblock \showarticletitle{Image analogies}. In
  \bibinfo{booktitle}{\emph{Proceedings of the 28th annual conference on
  Computer graphics and interactive techniques}}. \bibinfo{pages}{327--340}.
\newblock


\bibitem[\protect\citeauthoryear{H{\"{o}}llein, Johnson, and
  Nie{\ss}ner}{H{\"{o}}llein et~al\mbox{.}}{2021}]%
        {stylemesh}
\bibfield{author}{\bibinfo{person}{Lukas H{\"{o}}llein},
  \bibinfo{person}{Justin Johnson}, {and} \bibinfo{person}{Matthias
  Nie{\ss}ner}.} \bibinfo{year}{2021}\natexlab{}.
\newblock \showarticletitle{StyleMesh: Style Transfer for Indoor 3D Scene
  Reconstructions}.
\newblock \bibinfo{journal}{\emph{CoRR}}  \bibinfo{volume}{abs/2112.01530}
  (\bibinfo{year}{2021}).
\newblock
\showeprint[arXiv]{2112.01530}


\bibitem[\protect\citeauthoryear{Huang, Wang, Luo, Ma, Jiang, Zhu, Li, and
  Liu}{Huang et~al\mbox{.}}{2017}]%
        {8100228}
\bibfield{author}{\bibinfo{person}{Haozhi Huang}, \bibinfo{person}{Hao Wang},
  \bibinfo{person}{Wenhan Luo}, \bibinfo{person}{Lin Ma},
  \bibinfo{person}{Wenhao Jiang}, \bibinfo{person}{Xiaolong Zhu},
  \bibinfo{person}{Zhifeng Li}, {and} \bibinfo{person}{Wei Liu}.}
  \bibinfo{year}{2017}\natexlab{}.
\newblock \showarticletitle{Real-Time Neural Style Transfer for Videos}. In
  \bibinfo{booktitle}{\emph{Proceedings of the IEEE/CVF Conference on Computer
  Vision and Pattern Recognition}}. \bibinfo{pages}{7044--7052}.
\newblock


\bibitem[\protect\citeauthoryear{Huang, Tseng, Saini, Singh, and Yang}{Huang
  et~al\mbox{.}}{2021}]%
        {huang_2021_3d_scene_stylization}
\bibfield{author}{\bibinfo{person}{Hsin-Ping Huang}, \bibinfo{person}{Hung-Yu
  Tseng}, \bibinfo{person}{Saurabh Saini}, \bibinfo{person}{Maneesh Singh},
  {and} \bibinfo{person}{Ming-Hsuan Yang}.} \bibinfo{year}{2021}\natexlab{}.
\newblock \showarticletitle{Learning To Stylize Novel Views}.
\newblock  (\bibinfo{date}{October} \bibinfo{year}{2021}),
  \bibinfo{pages}{13869--13878}.
\newblock


\bibitem[\protect\citeauthoryear{Huang and Belongie}{Huang and
  Belongie}{2017}]%
        {huang2017adain}
\bibfield{author}{\bibinfo{person}{Xun Huang} {and} \bibinfo{person}{Serge
  Belongie}.} \bibinfo{year}{2017}\natexlab{}.
\newblock \showarticletitle{Arbitrary Style Transfer in Real-Time With Adaptive
  Instance Normalization}. In \bibinfo{booktitle}{\emph{Proceedings of the IEEE
  International Conference on Computer Vision}}.
\newblock


\bibitem[\protect\citeauthoryear{Ilg, Mayer, Saikia, Keuper, Dosovitskiy, and
  Brox}{Ilg et~al\mbox{.}}{2017}]%
        {flownet}
\bibfield{author}{\bibinfo{person}{E. Ilg}, \bibinfo{person}{N. Mayer},
  \bibinfo{person}{T. Saikia}, \bibinfo{person}{M. Keuper}, \bibinfo{person}{A.
  Dosovitskiy}, {and} \bibinfo{person}{T. Brox}.}
  \bibinfo{year}{2017}\natexlab{}.
\newblock \showarticletitle{FlowNet 2.0: Evolution of Optical Flow Estimation
  with Deep Networks}. In \bibinfo{booktitle}{\emph{Proceedings of the IEEE/CVF
  Conference on Computer Vision and Pattern Recognition}}.
\newblock


\bibitem[\protect\citeauthoryear{Jamri\v{s}ka, \v{S}\'{a}rka Sochorov\'{a},
  Texler, Luk\'{a}\v{c}, Fi\v{s}er, Lu, Shechtman, and S\'{y}kora}{Jamri\v{s}ka
  et~al\mbox{.}}{2019}]%
        {Jamriska19-SIG}
\bibfield{author}{\bibinfo{person}{Ond\v{r}ej Jamri\v{s}ka},
  \bibinfo{person}{\v{S}\'{a}rka Sochorov\'{a}}, \bibinfo{person}{Ond\v{r}ej
  Texler}, \bibinfo{person}{Michal Luk\'{a}\v{c}}, \bibinfo{person}{Jakub
  Fi\v{s}er}, \bibinfo{person}{Jingwan Lu}, \bibinfo{person}{Eli Shechtman},
  {and} \bibinfo{person}{Daniel S\'{y}kora}.} \bibinfo{year}{2019}\natexlab{}.
\newblock \showarticletitle{Stylizing Video by Example}.
\newblock \bibinfo{journal}{\emph{ACM Transactions on Graphics}}
  \bibinfo{volume}{38}, \bibinfo{number}{4}, Article \bibinfo{articleno}{107}
  (\bibinfo{year}{2019}).
\newblock


\bibitem[\protect\citeauthoryear{Johnson, Alahi, and Fei-Fei}{Johnson
  et~al\mbox{.}}{2016}]%
        {Johnson2016Perceptual}
\bibfield{author}{\bibinfo{person}{Justin Johnson}, \bibinfo{person}{Alexandre
  Alahi}, {and} \bibinfo{person}{Li Fei-Fei}.} \bibinfo{year}{2016}\natexlab{}.
\newblock \showarticletitle{Perceptual losses for real-time style transfer and
  super-resolution}. In \bibinfo{booktitle}{\emph{European Conference on
  Computer Vision}}.
\newblock


\bibitem[\protect\citeauthoryear{Kalantari, Wang, and Ramamoorthi}{Kalantari
  et~al\mbox{.}}{2016}]%
        {Kalantari16}
\bibfield{author}{\bibinfo{person}{Nima~Khademi Kalantari},
  \bibinfo{person}{Ting-Chun Wang}, {and} \bibinfo{person}{Ravi Ramamoorthi}.}
  \bibinfo{year}{2016}\natexlab{}.
\newblock \showarticletitle{Learning-Based View Synthesis for Light Field
  Cameras}.
\newblock \bibinfo{journal}{\emph{ACM Trans. Graph.}} \bibinfo{volume}{35},
  \bibinfo{number}{6}, Article \bibinfo{articleno}{193} (\bibinfo{date}{nov}
  \bibinfo{year}{2016}).
\newblock


\bibitem[\protect\citeauthoryear{Karras, Laine, and Aila}{Karras
  et~al\mbox{.}}{2019}]%
        {karras2019}
\bibfield{author}{\bibinfo{person}{Tero Karras}, \bibinfo{person}{Samuli
  Laine}, {and} \bibinfo{person}{Timo Aila}.} \bibinfo{year}{2019}\natexlab{}.
\newblock \showarticletitle{A Style-Based Generator Architecture for Generative
  Adversarial Networks}. In \bibinfo{booktitle}{\emph{Proceedings of the
  IEEE/CVF Conference on Computer Vision and Pattern Recognition}}.
\newblock


\bibitem[\protect\citeauthoryear{Kato, Ushiku, and Harada}{Kato
  et~al\mbox{.}}{2018}]%
        {kato2018renderer}
\bibfield{author}{\bibinfo{person}{Hiroharu Kato}, \bibinfo{person}{Yoshitaka
  Ushiku}, {and} \bibinfo{person}{Tatsuya Harada}.}
  \bibinfo{year}{2018}\natexlab{}.
\newblock \showarticletitle{Neural 3D Mesh Renderer}. In
  \bibinfo{booktitle}{\emph{Proceedings of the IEEE/CVF Conference on Computer
  Vision and Pattern Recognition}}.
\newblock


\bibitem[\protect\citeauthoryear{Kim, Kolkin, Salavon, and Shakhnarovich}{Kim
  et~al\mbox{.}}{2020}]%
        {Kim20DST}
\bibfield{author}{\bibinfo{person}{Sunnie S.~Y. Kim}, \bibinfo{person}{Nicholas
  Kolkin}, \bibinfo{person}{Jason Salavon}, {and} \bibinfo{person}{Gregory
  Shakhnarovich}.} \bibinfo{year}{2020}\natexlab{}.
\newblock \showarticletitle{Deformable Style Transfer}. In
  \bibinfo{booktitle}{\emph{Proceedings of the European Conference on Computer
  Vision}}.
\newblock


\bibitem[\protect\citeauthoryear{Knapitsch, Park, Zhou, and Koltun}{Knapitsch
  et~al\mbox{.}}{2017}]%
        {TankAndTemple}
\bibfield{author}{\bibinfo{person}{Arno Knapitsch}, \bibinfo{person}{Jaesik
  Park}, \bibinfo{person}{Qian-Yi Zhou}, {and} \bibinfo{person}{Vladlen
  Koltun}.} \bibinfo{year}{2017}\natexlab{}.
\newblock \showarticletitle{Tanks and Temples: Benchmarking Large-Scale Scene
  Reconstruction}.
\newblock \bibinfo{journal}{\emph{ACM Trans. Graph.}} \bibinfo{volume}{36},
  \bibinfo{number}{4}, Article \bibinfo{articleno}{78} (\bibinfo{date}{jul}
  \bibinfo{year}{2017}), \bibinfo{numpages}{13}~pages.
\newblock
\showISSN{0730-0301}


\bibitem[\protect\citeauthoryear{Kopanas, Philip, Leimk{\"u}hler, and
  Drettakis}{Kopanas et~al\mbox{.}}{2021}]%
        {KPLD21}
\bibfield{author}{\bibinfo{person}{Georgios Kopanas}, \bibinfo{person}{Julien
  Philip}, \bibinfo{person}{Thomas Leimk{\"u}hler}, {and}
  \bibinfo{person}{George Drettakis}.} \bibinfo{year}{2021}\natexlab{}.
\newblock \showarticletitle{Point-Based Neural Rendering with Per-View
  Optimization}.
\newblock \bibinfo{journal}{\emph{Computer Graphics Forum (Proceedings of the
  Eurographics Symposium on Rendering)}} \bibinfo{volume}{40},
  \bibinfo{number}{4} (\bibinfo{date}{June} \bibinfo{year}{2021}).
\newblock


\bibitem[\protect\citeauthoryear{Lai, Huang, Wang, Shechtman, Yumer, and
  Yang}{Lai et~al\mbox{.}}{2018}]%
        {Lai-ECCV-2018}
\bibfield{author}{\bibinfo{person}{Wei-Sheng Lai}, \bibinfo{person}{Jia-Bin
  Huang}, \bibinfo{person}{Oliver Wang}, \bibinfo{person}{Eli Shechtman},
  \bibinfo{person}{Ersin Yumer}, {and} \bibinfo{person}{Ming-Hsuan Yang}.}
  \bibinfo{year}{2018}\natexlab{}.
\newblock \showarticletitle{Learning Blind Video Temporal Consistency}. In
  \bibinfo{booktitle}{\emph{Proceedings of the European Conference on Computer
  Vision}}. \bibinfo{pages}{179–--195}.
\newblock


\bibitem[\protect\citeauthoryear{Levoy and Hanrahan}{Levoy and
  Hanrahan}{1996}]%
        {Levoy:96}
\bibfield{author}{\bibinfo{person}{Marc Levoy} {and} \bibinfo{person}{Pat
  Hanrahan}.} \bibinfo{year}{1996}\natexlab{}.
\newblock \showarticletitle{Light Field Rendering}. In
  \bibinfo{booktitle}{\emph{Proceedings of the 23rd Annual Conference on
  Computer Graphics and Interactive Techniques}}. \bibinfo{pages}{31–42}.
\newblock


\bibitem[\protect\citeauthoryear{Li, Liu, Kautz, and Yang}{Li
  et~al\mbox{.}}{2019}]%
        {li2018learning}
\bibfield{author}{\bibinfo{person}{Xueting Li}, \bibinfo{person}{Sifei Liu},
  \bibinfo{person}{Jan Kautz}, {and} \bibinfo{person}{Ming-Hsuan Yang}.}
  \bibinfo{year}{2019}\natexlab{}.
\newblock \showarticletitle{Learning Linear Transformations for Fast Image and
  Video Style Transfer}. In \bibinfo{booktitle}{\emph{Proceedings of the
  IEEE/CVF Conference on Computer Vision and Pattern Recognition}}.
\newblock


\bibitem[\protect\citeauthoryear{Li, Fang, Yang, Wang, Lu, and Yang}{Li
  et~al\mbox{.}}{2017}]%
        {WCT-NIPS-2017}
\bibfield{author}{\bibinfo{person}{Yijun Li}, \bibinfo{person}{Chen Fang},
  \bibinfo{person}{Jimei Yang}, \bibinfo{person}{Zhaowen Wang},
  \bibinfo{person}{Xin Lu}, {and} \bibinfo{person}{Ming-Hsuan Yang}.}
  \bibinfo{year}{2017}\natexlab{}.
\newblock \showarticletitle{Universal Style Transfer via Feature Transforms}.
  In \bibinfo{booktitle}{\emph{Advances in Neural Information Processing
  Systems}}.
\newblock


\bibitem[\protect\citeauthoryear{Li, Niklaus, Snavely, and Wang}{Li
  et~al\mbox{.}}{2021}]%
        {li2021neural}
\bibfield{author}{\bibinfo{person}{Zhengqi Li}, \bibinfo{person}{Simon
  Niklaus}, \bibinfo{person}{Noah Snavely}, {and} \bibinfo{person}{Oliver
  Wang}.} \bibinfo{year}{2021}\natexlab{}.
\newblock \showarticletitle{Neural scene flow fields for space-time view
  synthesis of dynamic scenes}. In \bibinfo{booktitle}{\emph{Proceedings of the
  IEEE/CVF Conference on Computer Vision and Pattern Recognition}}.
  \bibinfo{pages}{6498--6508}.
\newblock


\bibitem[\protect\citeauthoryear{Liao, Yao, Yuan, Hua, and Kang}{Liao
  et~al\mbox{.}}{2017}]%
        {deepImageAnalogies}
\bibfield{author}{\bibinfo{person}{Jing Liao}, \bibinfo{person}{Yuan Yao},
  \bibinfo{person}{Lu Yuan}, \bibinfo{person}{Gang Hua}, {and}
  \bibinfo{person}{Sing~Bing Kang}.} \bibinfo{year}{2017}\natexlab{}.
\newblock \showarticletitle{Visual Attribute Transfer through Deep Image
  Analogy}.
\newblock \bibinfo{journal}{\emph{ACM Trans. Graph.}} \bibinfo{volume}{36},
  \bibinfo{number}{4}, Article \bibinfo{articleno}{120} (\bibinfo{date}{jul}
  \bibinfo{year}{2017}), \bibinfo{numpages}{15}~pages.
\newblock
\showISSN{0730-0301}


\bibitem[\protect\citeauthoryear{Liu, Cheng, Lai, and Rosin}{Liu
  et~al\mbox{.}}{2017}]%
        {liu2017}
\bibfield{author}{\bibinfo{person}{Xiao-Chang Liu}, \bibinfo{person}{Ming-Ming
  Cheng}, \bibinfo{person}{Yu-Kun Lai}, {and} \bibinfo{person}{Paul~L. Rosin}.}
  \bibinfo{year}{2017}\natexlab{}.
\newblock \showarticletitle{Depth-Aware Neural Style Transfer}. In
  \bibinfo{booktitle}{\emph{Proceedings of the Symposium on Non-Photorealistic
  Animation and Rendering}} (Los Angeles, California)
  \emph{(\bibinfo{series}{NPAR '17})}. \bibinfo{publisher}{Association for
  Computing Machinery}, \bibinfo{address}{New York, NY, USA}, Article
  \bibinfo{articleno}{4}, \bibinfo{numpages}{10}~pages.
\newblock
\showISBNx{9781450350815}


\bibitem[\protect\citeauthoryear{Liu, Yang, and Hall}{Liu
  et~al\mbox{.}}{2021}]%
        {9577906}
\bibfield{author}{\bibinfo{person}{Xiao-Chang Liu}, \bibinfo{person}{Yong-Liang
  Yang}, {and} \bibinfo{person}{Peter Hall}.} \bibinfo{year}{2021}\natexlab{}.
\newblock \showarticletitle{Learning to Warp for Style Transfer}. In
  \bibinfo{booktitle}{\emph{Proceedings of the IEEE/CVF Conference on Computer
  Vision and Pattern Recognition}}. \bibinfo{pages}{3701--3710}.
\newblock


\bibitem[\protect\citeauthoryear{Ma, Huang, Sheffer, Kalogerakis, and Wang}{Ma
  et~al\mbox{.}}{2014}]%
        {Ma:2014:AST}
\bibfield{author}{\bibinfo{person}{Chongyang Ma}, \bibinfo{person}{Haibin
  Huang}, \bibinfo{person}{Alla Sheffer}, \bibinfo{person}{Evangelos
  Kalogerakis}, {and} \bibinfo{person}{Rui Wang}.}
  \bibinfo{year}{2014}\natexlab{}.
\newblock \showarticletitle{Analogy-Driven {3D} Style Transfer}.
\newblock \bibinfo{journal}{\emph{Computer Graphics Forum}}
  \bibinfo{volume}{33}, \bibinfo{number}{2} (\bibinfo{year}{2014}),
  \bibinfo{pages}{175--184}.
\newblock


\bibitem[\protect\citeauthoryear{Martin-Brualla, Radwan, Sajjadi, Barron,
  Dosovitskiy, and Duckworth}{Martin-Brualla et~al\mbox{.}}{2021}]%
        {martinbrualla2020nerfw}
\bibfield{author}{\bibinfo{person}{Ricardo Martin-Brualla},
  \bibinfo{person}{Noha Radwan}, \bibinfo{person}{Mehdi S.~M. Sajjadi},
  \bibinfo{person}{Jonathan~T. Barron}, \bibinfo{person}{Alexey Dosovitskiy},
  {and} \bibinfo{person}{Daniel Duckworth}.} \bibinfo{year}{2021}\natexlab{}.
\newblock \showarticletitle{NeRF in the Wild: Neural Radiance Fields for
  Unconstrained Photo Collections}. In \bibinfo{booktitle}{\emph{Proceedings of
  the IEEE/CVF Conference on Computer Vision and Pattern Recognition (CVPR)}}.
  \bibinfo{pages}{7210--7219}.
\newblock


\bibitem[\protect\citeauthoryear{Mildenhall, Srinivasan, Ortiz-Cayon,
  Kalantari, Ramamoorthi, Ng, and Kar}{Mildenhall et~al\mbox{.}}{2019a}]%
        {mildenhall2019local}
\bibfield{author}{\bibinfo{person}{Ben Mildenhall}, \bibinfo{person}{Pratul~P
  Srinivasan}, \bibinfo{person}{Rodrigo Ortiz-Cayon},
  \bibinfo{person}{Nima~Khademi Kalantari}, \bibinfo{person}{Ravi Ramamoorthi},
  \bibinfo{person}{Ren Ng}, {and} \bibinfo{person}{Abhishek Kar}.}
  \bibinfo{year}{2019}\natexlab{a}.
\newblock \showarticletitle{Local light field fusion: Practical view synthesis
  with prescriptive sampling guidelines}.
\newblock \bibinfo{journal}{\emph{ACM Transactions on Graphics (TOG)}}
  \bibinfo{volume}{38}, \bibinfo{number}{4} (\bibinfo{year}{2019}),
  \bibinfo{pages}{29:1--29:14}.
\newblock


\bibitem[\protect\citeauthoryear{Mildenhall, Srinivasan, Ortiz-Cayon,
  Kalantari, Ramamoorthi, Ng, and Kar}{Mildenhall et~al\mbox{.}}{2019b}]%
        {mildenhall2019llff}
\bibfield{author}{\bibinfo{person}{Ben Mildenhall}, \bibinfo{person}{Pratul~P.
  Srinivasan}, \bibinfo{person}{Rodrigo Ortiz-Cayon},
  \bibinfo{person}{Nima~Khademi Kalantari}, \bibinfo{person}{Ravi Ramamoorthi},
  \bibinfo{person}{Ren Ng}, {and} \bibinfo{person}{Abhishek Kar}.}
  \bibinfo{year}{2019}\natexlab{b}.
\newblock \showarticletitle{Local Light Field Fusion: Practical View Synthesis
  with Prescriptive Sampling Guidelines}.
\newblock \bibinfo{journal}{\emph{ACM Transactions on Graphics (TOG)}}
  (\bibinfo{year}{2019}).
\newblock


\bibitem[\protect\citeauthoryear{Mildenhall, Srinivasan, Tancik, Barron,
  Ramamoorthi, and Ng}{Mildenhall et~al\mbox{.}}{2020}]%
        {mildenhall2020nerf}
\bibfield{author}{\bibinfo{person}{Ben Mildenhall}, \bibinfo{person}{Pratul~P.
  Srinivasan}, \bibinfo{person}{Matthew Tancik}, \bibinfo{person}{Jonathan~T.
  Barron}, \bibinfo{person}{Ravi Ramamoorthi}, {and} \bibinfo{person}{Ren Ng}.}
  \bibinfo{year}{2020}\natexlab{}.
\newblock \showarticletitle{NeRF: Representing Scenes as Neural Radiance Fields
  for View Synthesis}. In \bibinfo{booktitle}{\emph{Proceedings of the European
  Conference on Computer Vision}}. \bibinfo{pages}{405--421}.
\newblock


\bibitem[\protect\citeauthoryear{M\"uller, Evans, Schied, and Keller}{M\"uller
  et~al\mbox{.}}{2022}]%
        {mueller2022instant}
\bibfield{author}{\bibinfo{person}{Thomas M\"uller}, \bibinfo{person}{Alex
  Evans}, \bibinfo{person}{Christoph Schied}, {and} \bibinfo{person}{Alexander
  Keller}.} \bibinfo{year}{2022}\natexlab{}.
\newblock \showarticletitle{Instant Neural Graphics Primitives with a
  Multiresolution Hash Encoding}.
\newblock \bibinfo{journal}{\emph{arXiv:2201.05989}} (\bibinfo{date}{Jan.}
  \bibinfo{year}{2022}).
\newblock


\bibitem[\protect\citeauthoryear{Neff, Stadlbauer, Parger, Kurz, Mueller,
  Chaitanya, Kaplanyan, and Steinberger}{Neff et~al\mbox{.}}{2021}]%
        {neff2021donerf}
\bibfield{author}{\bibinfo{person}{Thomas Neff}, \bibinfo{person}{Pascal
  Stadlbauer}, \bibinfo{person}{Mathias Parger}, \bibinfo{person}{Andreas
  Kurz}, \bibinfo{person}{Joerg~H. Mueller}, \bibinfo{person}{Chakravarty
  R.~Alla Chaitanya}, \bibinfo{person}{Anton~S. Kaplanyan}, {and}
  \bibinfo{person}{Markus Steinberger}.} \bibinfo{year}{2021}\natexlab{}.
\newblock \showarticletitle{{DONeRF: Towards Real-Time Rendering of Compact
  Neural Radiance Fields using Depth Oracle Networks}}.
\newblock \bibinfo{journal}{\emph{Computer Graphics Forum}}
  \bibinfo{volume}{40}, \bibinfo{number}{4} (\bibinfo{year}{2021}).
\newblock
\showISSN{1467-8659}


\bibitem[\protect\citeauthoryear{Nguyen, Ritschel, Myszkowski, Eisemann, and
  Seidel}{Nguyen et~al\mbox{.}}{2012}]%
        {nguyen:2012:3DMaterialStyle}
\bibfield{author}{\bibinfo{person}{Chuong~H. Nguyen}, \bibinfo{person}{Tobias
  Ritschel}, \bibinfo{person}{Karol Myszkowski}, \bibinfo{person}{Elmar
  Eisemann}, {and} \bibinfo{person}{Hans-Peter Seidel}.}
  \bibinfo{year}{2012}\natexlab{}.
\newblock \showarticletitle{{3D Material Style Transfer}}.
\newblock \bibinfo{journal}{\emph{Computer Graphics Forum (Proc. EUROGRAPHICS
  2012)}} \bibinfo{volume}{2}, \bibinfo{number}{31} (\bibinfo{year}{2012}).
\newblock


\bibitem[\protect\citeauthoryear{Penner and Zhang}{Penner and Zhang}{2017}]%
        {penner2017soft}
\bibfield{author}{\bibinfo{person}{Eric Penner} {and} \bibinfo{person}{Li
  Zhang}.} \bibinfo{year}{2017}\natexlab{}.
\newblock \showarticletitle{Soft 3D reconstruction for view synthesis}.
\newblock \bibinfo{journal}{\emph{ACM Transactions on Graphics (TOG)}}
  \bibinfo{volume}{36}, \bibinfo{number}{6} (\bibinfo{year}{2017}),
  \bibinfo{pages}{1--11}.
\newblock


\bibitem[\protect\citeauthoryear{Radford, Kim, Hallacy, Ramesh, Goh, Agarwal,
  Sastry, Askell, Mishkin, Clark, Krueger, and Sutskever}{Radford
  et~al\mbox{.}}{2021}]%
        {CLIP}
\bibfield{author}{\bibinfo{person}{Alec Radford}, \bibinfo{person}{Jong~Wook
  Kim}, \bibinfo{person}{Chris Hallacy}, \bibinfo{person}{Aditya Ramesh},
  \bibinfo{person}{Gabriel Goh}, \bibinfo{person}{Sandhini Agarwal},
  \bibinfo{person}{Girish Sastry}, \bibinfo{person}{Amanda Askell},
  \bibinfo{person}{Pamela Mishkin}, \bibinfo{person}{Jack Clark},
  \bibinfo{person}{Gretchen Krueger}, {and} \bibinfo{person}{Ilya Sutskever}.}
  \bibinfo{year}{2021}\natexlab{}.
\newblock \showarticletitle{Learning Transferable Visual Models From Natural
  Language Supervision}. In \bibinfo{booktitle}{\emph{ICML}}
  \emph{(\bibinfo{series}{Proceedings of Machine Learning Research},
  Vol.~\bibinfo{volume}{139})}, \bibfield{editor}{\bibinfo{person}{Marina
  Meila} {and} \bibinfo{person}{Tong Zhang}} (Eds.).
  \bibinfo{pages}{8748--8763}.
\newblock


\bibitem[\protect\citeauthoryear{Segu, Grinvald, Siegwart, and Tombari}{Segu
  et~al\mbox{.}}{2020}]%
        {segu20203dsnet}
\bibfield{author}{\bibinfo{person}{Mattia Segu}, \bibinfo{person}{Margarita
  Grinvald}, \bibinfo{person}{Roland Siegwart}, {and} \bibinfo{person}{Federico
  Tombari}.} \bibinfo{year}{2020}\natexlab{}.
\newblock \showarticletitle{3DSNet: Unsupervised Shape-to-Shape 3D Style
  Transfer}.
\newblock \bibinfo{journal}{\emph{arXiv preprint arXiv:2011.13388}}
  (\bibinfo{year}{2020}).
\newblock


\bibitem[\protect\citeauthoryear{Simonyan and Zisserman}{Simonyan and
  Zisserman}{2015}]%
        {VGG}
\bibfield{author}{\bibinfo{person}{Karen Simonyan} {and}
  \bibinfo{person}{Andrew Zisserman}.} \bibinfo{year}{2015}\natexlab{}.
\newblock \showarticletitle{Very Deep Convolutional Networks for Large-Scale
  Image Recognition}. In \bibinfo{booktitle}{\emph{3rd International Conference
  on Learning Representations, {ICLR} 2015, San Diego, CA, USA, May 7-9, 2015,
  Conference Track Proceedings}}, \bibfield{editor}{\bibinfo{person}{Yoshua
  Bengio} {and} \bibinfo{person}{Yann LeCun}} (Eds.).
\newblock


\bibitem[\protect\citeauthoryear{Sitzmann, Thies, Heide, Nie{\ss}ner,
  Wetzstein, and Zollhofer}{Sitzmann et~al\mbox{.}}{2019}]%
        {sitzmann2019deepvoxels}
\bibfield{author}{\bibinfo{person}{Vincent Sitzmann}, \bibinfo{person}{Justus
  Thies}, \bibinfo{person}{Felix Heide}, \bibinfo{person}{Matthias
  Nie{\ss}ner}, \bibinfo{person}{Gordon Wetzstein}, {and}
  \bibinfo{person}{Michael Zollhofer}.} \bibinfo{year}{2019}\natexlab{}.
\newblock \showarticletitle{Deepvoxels: Learning persistent 3d feature
  embeddings}. In \bibinfo{booktitle}{\emph{Proceedings of the IEEE/CVF
  Conference on Computer Vision and Pattern Recognition}}.
  \bibinfo{pages}{2437--2446}.
\newblock


\bibitem[\protect\citeauthoryear{Spirin}{Spirin}{2021}]%
        {arcanegan}
\bibfield{author}{\bibinfo{person}{Alex Spirin}.}
  \bibinfo{year}{2021}\natexlab{}.
\newblock \bibinfo{booktitle}{\emph{ArcaneGAN}}.
\newblock
\urldef\tempurl%
\url{https://github.com/Sxela/ArcaneGAN}
\showURL{%
\tempurl}


\bibitem[\protect\citeauthoryear{Srinivasan, Tucker, Barron, Ramamoorthi, Ng,
  and Snavely}{Srinivasan et~al\mbox{.}}{2019}]%
        {srinivasan2019pushing}
\bibfield{author}{\bibinfo{person}{Pratul~P Srinivasan},
  \bibinfo{person}{Richard Tucker}, \bibinfo{person}{Jonathan~T Barron},
  \bibinfo{person}{Ravi Ramamoorthi}, \bibinfo{person}{Ren Ng}, {and}
  \bibinfo{person}{Noah Snavely}.} \bibinfo{year}{2019}\natexlab{}.
\newblock \showarticletitle{Pushing the boundaries of view extrapolation with
  multiplane images}. In \bibinfo{booktitle}{\emph{Proceedings of the IEEE/CVF
  Conference on Computer Vision and Pattern Recognition}}.
  \bibinfo{pages}{175--184}.
\newblock


\bibitem[\protect\citeauthoryear{Svoboda, Anoosheh, Osendorfer, and
  Masci}{Svoboda et~al\mbox{.}}{2020}]%
        {svoboda2020twostage}
\bibfield{author}{\bibinfo{person}{Jan Svoboda}, \bibinfo{person}{Asha
  Anoosheh}, \bibinfo{person}{Christian Osendorfer}, {and}
  \bibinfo{person}{Jonathan Masci}.} \bibinfo{year}{2020}\natexlab{}.
\newblock \showarticletitle{Two-Stage Peer-Regularized Feature Recombination
  for Arbitrary Image Style Transfer}. In \bibinfo{booktitle}{\emph{Proceedings
  of the IEEE/CVF Conference on Computer Vision and Pattern Recognition}}.
\newblock


\bibitem[\protect\citeauthoryear{S\'{y}kora, Jamri\v{s}ka, Texler, Fi\v{s}er,
  Luk\'{a}\v{c}, Lu, and Shechtman}{S\'{y}kora et~al\mbox{.}}{2019a}]%
        {styblit}
\bibfield{author}{\bibinfo{person}{Daniel S\'{y}kora},
  \bibinfo{person}{Ond\v{r}ej Jamri\v{s}ka}, \bibinfo{person}{Ond\v{r}ej
  Texler}, \bibinfo{person}{Jakub Fi\v{s}er}, \bibinfo{person}{Michal
  Luk\'{a}\v{c}}, \bibinfo{person}{Jingwan Lu}, {and} \bibinfo{person}{Eli
  Shechtman}.} \bibinfo{year}{2019}\natexlab{a}.
\newblock \showarticletitle{{StyleBlit}: Fast Example-Based Stylization with
  Local Guidance}.
\newblock \bibinfo{journal}{\emph{Computer Graphics Forum}}
  \bibinfo{volume}{38}, \bibinfo{number}{2} (\bibinfo{year}{2019}),
  \bibinfo{pages}{83--91}.
\newblock


\bibitem[\protect\citeauthoryear{S\'{y}kora, Jamri\v{s}ka, Texler, Fi\v{s}er,
  Luk\'{a}\v{c}, Lu, and Shechtman}{S\'{y}kora et~al\mbox{.}}{2019b}]%
        {Sykora19-EG}
\bibfield{author}{\bibinfo{person}{Daniel S\'{y}kora},
  \bibinfo{person}{Ond\v{r}ej Jamri\v{s}ka}, \bibinfo{person}{Ond\v{r}ej
  Texler}, \bibinfo{person}{Jakub Fi\v{s}er}, \bibinfo{person}{Michal
  Luk\'{a}\v{c}}, \bibinfo{person}{Jingwan Lu}, {and} \bibinfo{person}{Eli
  Shechtman}.} \bibinfo{year}{2019}\natexlab{b}.
\newblock \showarticletitle{{StyleBlit}: Fast Example-Based Stylization with
  Local Guidance}.
\newblock \bibinfo{journal}{\emph{Computer Graphics Forum}}
  \bibinfo{volume}{38}, \bibinfo{number}{2} (\bibinfo{year}{2019}),
  \bibinfo{pages}{83--91}.
\newblock


\bibitem[\protect\citeauthoryear{Teed and Deng}{Teed and Deng}{2021}]%
        {RAFT}
\bibfield{author}{\bibinfo{person}{Zachary Teed} {and} \bibinfo{person}{Jia
  Deng}.} \bibinfo{year}{2021}\natexlab{}.
\newblock \showarticletitle{RAFT: Recurrent All-Pairs Field Transforms for
  Optical Flow (Extended Abstract)}. In \bibinfo{booktitle}{\emph{Proceedings
  of the Thirtieth International Joint Conference on Artificial Intelligence,
  {IJCAI-21}}}, \bibfield{editor}{\bibinfo{person}{Zhi-Hua Zhou}} (Ed.).
  \bibinfo{publisher}{International Joint Conferences on Artificial
  Intelligence Organization}, \bibinfo{pages}{4839--4843}.
\newblock


\bibitem[\protect\citeauthoryear{Texler, Fi\v{s}er, Luk\'{a}\v{c}, Lu,
  Shechtman, and S\'{y}kora}{Texler et~al\mbox{.}}{2019}]%
        {Texler19-NPAR}
\bibfield{author}{\bibinfo{person}{Ond\v{r}ej Texler}, \bibinfo{person}{Jakub
  Fi\v{s}er}, \bibinfo{person}{Michal Luk\'{a}\v{c}}, \bibinfo{person}{Jingwan
  Lu}, \bibinfo{person}{Eli Shechtman}, {and} \bibinfo{person}{Daniel
  S\'{y}kora}.} \bibinfo{year}{2019}\natexlab{}.
\newblock \showarticletitle{Enhancing Neural Style Transfer using Patch-Based
  Synthesis}. In \bibinfo{booktitle}{\emph{Proceedings of the 8th ACM/EG
  Expressive Symposium}}. \bibinfo{pages}{43--50}.
\newblock


\bibitem[\protect\citeauthoryear{Texler, Futschik, Ku\v{c}era, Jamri\v{s}ka,
  \v{S}\'{a}rka Sochorov\'{a}, Chai, Tulyakov, and S\'{y}kora}{Texler
  et~al\mbox{.}}{2020}]%
        {Texler20-SIG}
\bibfield{author}{\bibinfo{person}{Ond\v{r}ej Texler}, \bibinfo{person}{David
  Futschik}, \bibinfo{person}{Michal Ku\v{c}era}, \bibinfo{person}{Ond\v{r}ej
  Jamri\v{s}ka}, \bibinfo{person}{\v{S}\'{a}rka Sochorov\'{a}},
  \bibinfo{person}{Menglei Chai}, \bibinfo{person}{Sergey Tulyakov}, {and}
  \bibinfo{person}{Daniel S\'{y}kora}.} \bibinfo{year}{2020}\natexlab{}.
\newblock \showarticletitle{Interactive Video Stylization Using Few-Shot
  Patch-Based Training}.
\newblock \bibinfo{journal}{\emph{ACM Transactions on Graphics}}
  \bibinfo{volume}{39}, \bibinfo{number}{4} (\bibinfo{year}{2020}),
  \bibinfo{pages}{73}.
\newblock


\bibitem[\protect\citeauthoryear{Thies, Zollh{\"o}fer, and Nie{\ss}ner}{Thies
  et~al\mbox{.}}{2019}]%
        {thies2019deferred}
\bibfield{author}{\bibinfo{person}{Justus Thies}, \bibinfo{person}{Michael
  Zollh{\"o}fer}, {and} \bibinfo{person}{Matthias Nie{\ss}ner}.}
  \bibinfo{year}{2019}\natexlab{}.
\newblock \showarticletitle{Deferred neural rendering: Image synthesis using
  neural textures}.
\newblock \bibinfo{journal}{\emph{ACM Transactions on Graphics (TOG)}}
  \bibinfo{volume}{38}, \bibinfo{number}{4} (\bibinfo{year}{2019}),
  \bibinfo{pages}{1--12}.
\newblock


\bibitem[\protect\citeauthoryear{Ulyanov, Vedaldi, and Lempitsky}{Ulyanov
  et~al\mbox{.}}{2016}]%
        {instanceNorm}
\bibfield{author}{\bibinfo{person}{Dmitry Ulyanov}, \bibinfo{person}{Andrea
  Vedaldi}, {and} \bibinfo{person}{Victor~S. Lempitsky}.}
  \bibinfo{year}{2016}\natexlab{}.
\newblock \showarticletitle{Instance Normalization: The Missing Ingredient for
  Fast Stylization}.
\newblock \bibinfo{journal}{\emph{CoRR}}  \bibinfo{volume}{abs/1607.08022}
  (\bibinfo{year}{2016}).
\newblock
\showeprint[arXiv]{1607.08022}


\bibitem[\protect\citeauthoryear{Wang, Yang, Xu, and Liu}{Wang
  et~al\mbox{.}}{2020a}]%
        {ReReVST2020}
\bibfield{author}{\bibinfo{person}{Wenjing Wang}, \bibinfo{person}{Shuai Yang},
  \bibinfo{person}{Jizheng Xu}, {and} \bibinfo{person}{Jiaying Liu}.}
  \bibinfo{year}{2020}\natexlab{a}.
\newblock \showarticletitle{Consistent Video Style Transfer via Relaxation and
  Regularization}.
\newblock \bibinfo{journal}{\emph{{IEEE} Trans. Image Process.}}
  (\bibinfo{year}{2020}).
\newblock


\bibitem[\protect\citeauthoryear{Wang, Zhao, Chen, Zuo, Li, Xing, and Lu}{Wang
  et~al\mbox{.}}{2021}]%
        {Wang2021}
\bibfield{author}{\bibinfo{person}{Zhizhong Wang}, \bibinfo{person}{Lei Zhao},
  \bibinfo{person}{Haibo Chen}, \bibinfo{person}{Zhiwen Zuo},
  \bibinfo{person}{Ailin Li}, \bibinfo{person}{Wei Xing}, {and}
  \bibinfo{person}{Dongming Lu}.} \bibinfo{year}{2021}\natexlab{}.
\newblock \showarticletitle{Diversified Patch-based Style Transfer with Shifted
  Style Normalization}.
\newblock \bibinfo{journal}{\emph{CoRR}}  \bibinfo{volume}{abs/2101.06381}
  (\bibinfo{year}{2021}).
\newblock
\showeprint[arXiv]{2101.06381}


\bibitem[\protect\citeauthoryear{Wang, Zhao, Lin, Mo, Zhang, Xing, and Lu}{Wang
  et~al\mbox{.}}{2020b}]%
        {wang2020glstylenet}
\bibfield{author}{\bibinfo{person}{Zhizhong Wang}, \bibinfo{person}{Lei Zhao},
  \bibinfo{person}{Sihuan Lin}, \bibinfo{person}{Qihang Mo},
  \bibinfo{person}{Huiming Zhang}, \bibinfo{person}{Wei Xing}, {and}
  \bibinfo{person}{Dongming Lu}.} \bibinfo{year}{2020}\natexlab{b}.
\newblock \showarticletitle{GLStyleNet: exquisite style transfer combining
  global and local pyramid features}.
\newblock \bibinfo{journal}{\emph{IET Computer Vision}} \bibinfo{volume}{14},
  \bibinfo{number}{8} (\bibinfo{year}{2020}), \bibinfo{pages}{575--586}.
\newblock


\bibitem[\protect\citeauthoryear{Xia, Xue, Lai, Sun, Chang, Kulis, and
  Chen}{Xia et~al\mbox{.}}{2021}]%
        {9423312}
\bibfield{author}{\bibinfo{person}{Xide Xia}, \bibinfo{person}{Tianfan Xue},
  \bibinfo{person}{Wei-sheng Lai}, \bibinfo{person}{Zheng Sun},
  \bibinfo{person}{Abby Chang}, \bibinfo{person}{Brian Kulis}, {and}
  \bibinfo{person}{Jiawen Chen}.} \bibinfo{year}{2021}\natexlab{}.
\newblock \showarticletitle{Real-time Localized Photorealistic Video Style
  Transfer}. In \bibinfo{booktitle}{\emph{2021 IEEE Winter Conference on
  Applications of Computer Vision (WACV)}}. \bibinfo{pages}{1088--1097}.
\newblock


\bibitem[\protect\citeauthoryear{Yin, Gao, Shugrina, Khamis, and Fidler}{Yin
  et~al\mbox{.}}{2021}]%
        {yin2021_3DStyleNet}
\bibfield{author}{\bibinfo{person}{Kangxue Yin}, \bibinfo{person}{Jun Gao},
  \bibinfo{person}{Maria Shugrina}, \bibinfo{person}{Sameh Khamis}, {and}
  \bibinfo{person}{Sanja Fidler}.} \bibinfo{year}{2021}\natexlab{}.
\newblock \showarticletitle{3DStyleNet: Creating 3D Shapes With Geometric and
  Texture Style Variations}. In \bibinfo{booktitle}{\emph{Proceedings of the
  IEEE/CVF International Conference on Computer Vision}}.
  \bibinfo{pages}{12456--12465}.
\newblock


\bibitem[\protect\citeauthoryear{Zhang, Riegler, Snavely, and Koltun}{Zhang
  et~al\mbox{.}}{2020}]%
        {kaizhang2020}
\bibfield{author}{\bibinfo{person}{Kai Zhang}, \bibinfo{person}{Gernot
  Riegler}, \bibinfo{person}{Noah Snavely}, {and} \bibinfo{person}{Vladlen
  Koltun}.} \bibinfo{year}{2020}\natexlab{}.
\newblock \showarticletitle{NeRF++: Analyzing and Improving Neural Radiance
  Fields}.
\newblock \bibinfo{journal}{\emph{arXiv:2010.07492}} (\bibinfo{year}{2020}).
\newblock


\bibitem[\protect\citeauthoryear{Zhou, Tucker, Flynn, Fyffe, and Snavely}{Zhou
  et~al\mbox{.}}{2018}]%
        {zhou2018stereo}
\bibfield{author}{\bibinfo{person}{Tinghui Zhou}, \bibinfo{person}{Richard
  Tucker}, \bibinfo{person}{John Flynn}, \bibinfo{person}{Graham Fyffe}, {and}
  \bibinfo{person}{Noah Snavely}.} \bibinfo{year}{2018}\natexlab{}.
\newblock \showarticletitle{Stereo magnification: learning view synthesis using
  multiplane images}.
\newblock \bibinfo{journal}{\emph{ACM Transactions on Graphics (TOG)}}
  \bibinfo{volume}{37}, \bibinfo{number}{4} (\bibinfo{year}{2018}),
  \bibinfo{pages}{1--12}.
\newblock


\bibitem[\protect\citeauthoryear{Zitnick, Kang, Uyttendaele, Winder, and
  Szeliski}{Zitnick et~al\mbox{.}}{2004}]%
        {zitnick2004high}
\bibfield{author}{\bibinfo{person}{C~Lawrence Zitnick},
  \bibinfo{person}{Sing~Bing Kang}, \bibinfo{person}{Matthew Uyttendaele},
  \bibinfo{person}{Simon Winder}, {and} \bibinfo{person}{Richard Szeliski}.}
  \bibinfo{year}{2004}\natexlab{}.
\newblock \showarticletitle{High-quality video view interpolation using a
  layered representation}.
\newblock \bibinfo{journal}{\emph{ACM transactions on graphics (TOG)}}
  \bibinfo{volume}{23}, \bibinfo{number}{3} (\bibinfo{year}{2004}),
  \bibinfo{pages}{600--608}.
\newblock


\end{thebibliography}
\end{document}